\pdfoutput=1

\documentclass[11pt]{article}
\usepackage{tabularx}
\usepackage[review]{ACL2023}
\usepackage{amssymb}

\usepackage{times}
\usepackage{latexsym}
\usepackage{graphicx}
\usepackage[T1]{fontenc}
\usepackage{algorithm, algorithmic}
\usepackage{varwidth}
\usepackage[utf8]{inputenc}

\usepackage{microtype}
\usepackage{amsbsy, textcomp, amsmath, caption, amsthm, subfigure}
\usepackage{inconsolata}
\usepackage{multirow,booktabs}
\usepackage{graphicx}
\usepackage{subcaption}
\usepackage{CJKutf8}

\usepackage{xcolor}

\definecolor{tealcolor}{RGB}{217, 234, 211}
\definecolor{lowred}{RGB}{255, 147, 147}

\title{Beyond Entity Alignment: Towards Complete Knowledge Graph Alignment via Entity-Relation  Synergy}

\author{Xiaohan Fang\thanks{\ \ School of Information Science and Engineering, Yanshan University.} \\
  \texttt{fangxiaohan98@stumail.ysu.edu.cn} \And
  Chaozhuo Li\thanks{\ \ Beijing University of Posts and Telecommunications.} \\
  \texttt{cli@microsoft.com} \And
  Yi Zhao \footnotemark[1] \\
  \texttt{zhaoyi.hb@gmail.com} \AND
  Qian Zang \footnotemark[1] \\
  \texttt{ysuzangqian@stumail.ysu.edu.cn}  \And
  Litian Zhang \thanks{\ \ Beihang University.} \\
  \texttt{litianzhang@buaa.edu.cn} \AND
  Jiquan Pen\footnotemark[1] \\
  \texttt{pengjiquan@stumail.ysu.edu.cn} \And
  Xi Zhang \footnotemark[2] \\
  \texttt{zhangx@bupt.edu.cn} \And
  Jibing Gong \thanks{\ \ Corresponding authors.} \footnotemark[1] \\
  \texttt{gongjibing@ysu.edu.cn}
}

\begin{document}
\maketitle
\begin{abstract}

Knowledge Graph Alignment (KGA) aims to integrate knowledge from multiple sources to address the limitations of individual Knowledge Graphs (KGs) in terms of coverage and depth. 
However, current KGA models fall short in achieving a ``complete'' knowledge graph alignment. 
Existing models primarily emphasize the linkage of cross-graph entities but overlook aligning relations across KGs, thereby providing only a partial solution to KGA.
The semantic correlations embedded in relations are largely overlooked, potentially restricting a comprehensive understanding of cross-KG signals. 
In this paper, we propose to conceptualize relation alignment as an independent task and conduct KGA by decomposing it into two distinct but highly correlated sub-tasks: entity alignment and relation alignment. To capture the mutually reinforcing correlations between these objectives, we propose a novel Expectation-Maximization-based model, EREM, which iteratively optimizes both sub-tasks. 
Experimental results on real-world datasets demonstrate that EREM consistently outperforms state-of-the-art models in both entity alignment and relation alignment tasks.

\end{abstract}

\section{Introduction}

Knowledge Graphs (KGs), structured as sets of triples (head entity, relation, and tail entity), function as a conceptual representation of factual knowledge extracted from real-world data. 
KGs commonly serve as a crucial reference and supplementary repository of knowledge across diverse domains \cite{wang2017knowledge, ji2021survey}. 
A solitary KG frequently lacks the breadth and depth of information required to adequately support various applications \cite{zhang2022benchmark}, suffering from the inherent challenges in comprehensively capturing and representing diverse knowledge domains within a single graph structure. 
Consequently, there arises a pressing need for strategies to address this insufficiency and enable the effective integration of knowledge from multiple sources, dubbed knowledge graph alignment (KGA).






Existing KGA models primarily concentrate on linking entities across different KGs \cite{li2019semi, sun2018bootstrapping, ijcai2019p733, liu-etal-2020-exploring, zeng2022interactive}. 
Initially, textual descriptions and relations are encoded into low-dimensional embeddings to represent entities. Subsequently, a few or none alignment seeds are employed to train a matching function that minimizes the distance between paired entities. 
The matching function deduces equivalent entities based on the entity embeddings, often formalized as a global assignment problem \cite{mao2021alignment} or an optimal transport problem \cite{tang2023fused, luo-yu-2022-accurate}.

\begin{figure}[t]%
\centering
\includegraphics[width=0.5\textwidth]{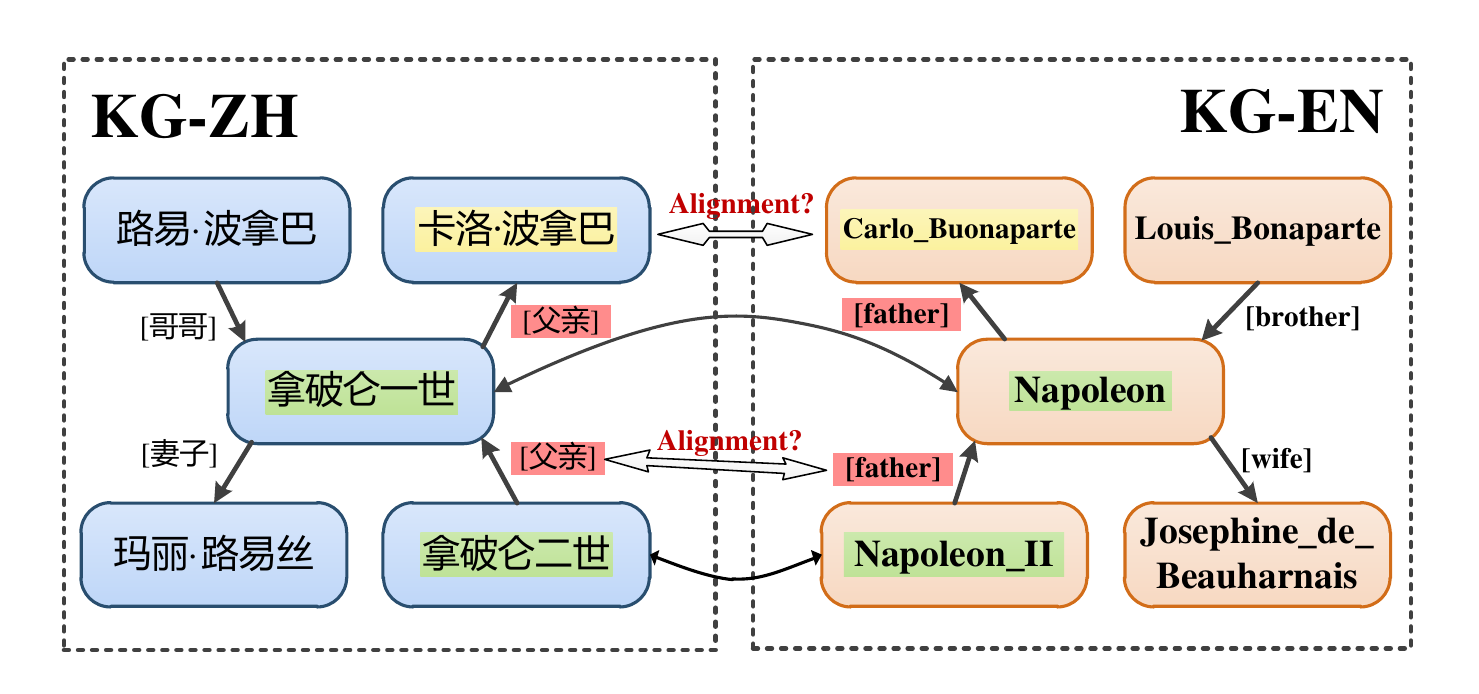}
\caption{An illustrative example demonstrates the mutually reinforcing correlations between EA and RA.}\label{fig2-1}
\end{figure}

Despite the advanced performance of existing approaches, they essentially focus on \textit{\textbf{entity alignment}} rather than \textit{\textbf{knowledge graph alignment}}. 
KGs consist of two fundamental components: entities and relations.  
Entities denote the intrinsic attributes of the corresponding objects, whereas relations elucidate the semantic correlations across entities.
Intuitively, an optimal KGA model should be capable of aligning both entities and relations within a unified framework. 
Nonetheless, existing research concentrates on entity alignment, often neglecting the alignment of relations. 
This deficiency results in a situation where only a partial aspect of the studied problem is addressed,  leading to inadequate or erroneous integration of knowledge sourced from diverse knowledge graphs.
Additionally, relations are only incorporated to improve the quality of entity embeddings, which overlooks significant cross-KG semantics inherent in relations \cite{zhang2022benchmark}. 
\begin{CJK*}{UTF8}{gbsn} In Figure \ref{fig2-1}, relation ``父亲'' in KG-ZH shares the identical semantics semantics the relation ``Father'' in KG-EN. \end{CJK*} 
Such semantic correlations embedded in relations have largely been disregarded,  potentially limiting the comprehensive understanding of cross-KG signals.

Distinct from existing entity alignment models, this paper aims to address the ``complete'' task of KGA. 
Our motivation lies in  formalizing KGA as two interrelated sub-tasks: Entity Alignment (EA) and Relation Alignment (RA). Similar to entity alignment, relation alignment seeks to match identical relations across different KGs. 
These two sub-tasks are interdependent and mutually reinforcing, creating a synergistic effect that enhances the overall alignment process.
On one hand, accurate entity alignment provides valuable context that significantly aids in the alignment of relations. 
\begin{CJK*}{UTF8}{gbsn} 
For instance, given the aligned entity pairs <``拿破仑一世'', ``Napoleon''> and <``拿破仑二世'', ``Napoleon\_II">, it is intuitive to infer that the relations connecting these entities (i.e., the KG-ZH relation ``父亲'' and the KG-EN relation ``father'') tend to share the identical meaning.
On the other hand,  precise alignment of relations can assist in reinforcing consistency constraints for entity alignment. 
For instance, if we establish that the head entity pair <``拿破仑一世'', ``Napoleon''> and the relation pair <``父亲'', ``father''> are aligned, it becomes straightforward to infer that the tail head entity pair <``卡洛·波拿巴'', ``Louis\_Bonaparte''> refers to the same individual. 
\end{CJK*}

While the EA-RA alignment task represents a comprehensive endeavor with promising co-evolution potential, it suffers from several challenges. 
First, designing a relation matching module to effectively align cross-KG relations remains non-trivial, particularly given the limited existing works. 
A straightforward approach involves aligning relations based on text similarity,  which neglects the structural characteristics of relations and the contextual importance of entities in relation matching \cite{zhu2021raga}. 
Second, designing an integrated framework for mutual iterative enhancement between EA and RA poses a significant challenge.   
Entity alignment necessitates leveraging multiple relations to achieve more robust entity representation. 
Conversely, relation alignment needs to consider the interconnected entities to effectively capture contextual nuances associated with relations. 
Balancing these requirements within a unified framework presents a complex task, requiring consideration of the distinct characteristics and objectives of entity and relation alignment.

To address the aforementioned challenges, we propose a novel integrated framework for mutual enhancement between EA and RA, dubbed \underline{EREM}. 
EREM consists of two modules: an Entity Matching (E-step) module and a Relation Matching (M-step) module, and is defined as a variational Expectation Maximization framework.
In the E-step, EA is optimized using the relation anchors learned by RA, with the objective of maximizing entity anchor correspondences. 
In the M-step, RA is optimized by leveraging the entity anchors predicted by EA, with the aim of maximizing correspondences between relation anchors. 
The process of matching EA and RA is formulated as an optimal transport (OT) alignment task, which is efficiently solved by the Sinkhorn algorithm\cite{cuturi2013sinkhorn}. 
Experimental results on several widely-used datasets demonstrate the superior performance of our proposed framework. 
Notably, seven SOTA EA methods can be seamlessly integrated into EREM, consistently exhibiting performance improvements.

The contributions of this paper are three folds:

\begin{itemize}

\item  To the best of our knowledge, we are the first to conceptualize relation alignment as an independent task. The ``complete'' knowledge graph alignment task is decomposed into two distinct but highly correlated sub-tasks.


\item To capture the mutually reinforcing correlations between the objectives of entity alignment and relation alignment, we propose EREM, an Expectation Maximization (EM) framework to iteratively optimize these two tasks, thereby achieving a unified solution.



\item We conduct extensive experiments on real-world datasets, demonstrating that our proposed framework consistently outperforms popular SOTA models in both the entity alignment and relation alignment tasks.

\end{itemize}

\section{Problem Definition}

\textbf{Definition 1. Knowledge Graph.} A knowledge graph is denoted as $G= \{E, R, T, S_e, S_r\}$, in which $ e \in E, r \in R, t = ( e_h, r , e_t ) \in T$ represents entity, relation and the triple (head entity, relation, tail entity),  respectively. 
$S_{e}$ includes the side information of entities such as name and description.
$S_{r}$ contains the  side information of relations like the associated name. 


\noindent \textbf{Definition 2. Entity Alignment (EA).} 
Assuming that the entities of two knowledge graphs follow distinct distributions, the entity alignment task aims to learn a transfer function to align the entities into a uniform distribution. 
Formally, given an entity $e$ in the source graph $G$, EA aims to find its equivalent entity $e'$ in the target graph $G'$.

\noindent \textbf{Definition 3. Relation Alignment (RA).}  
RA aims to learn a transfer function to align two KG's relation distributions into a uniform distribution.
For each relation type $r$ in the source graph, RA locates its equivalent relation  $r'$ in the target graph. 

\section{Methodology}

\begin{figure*}[h]

\begin{center}

\includegraphics[width=16cm,height=11.5cm]{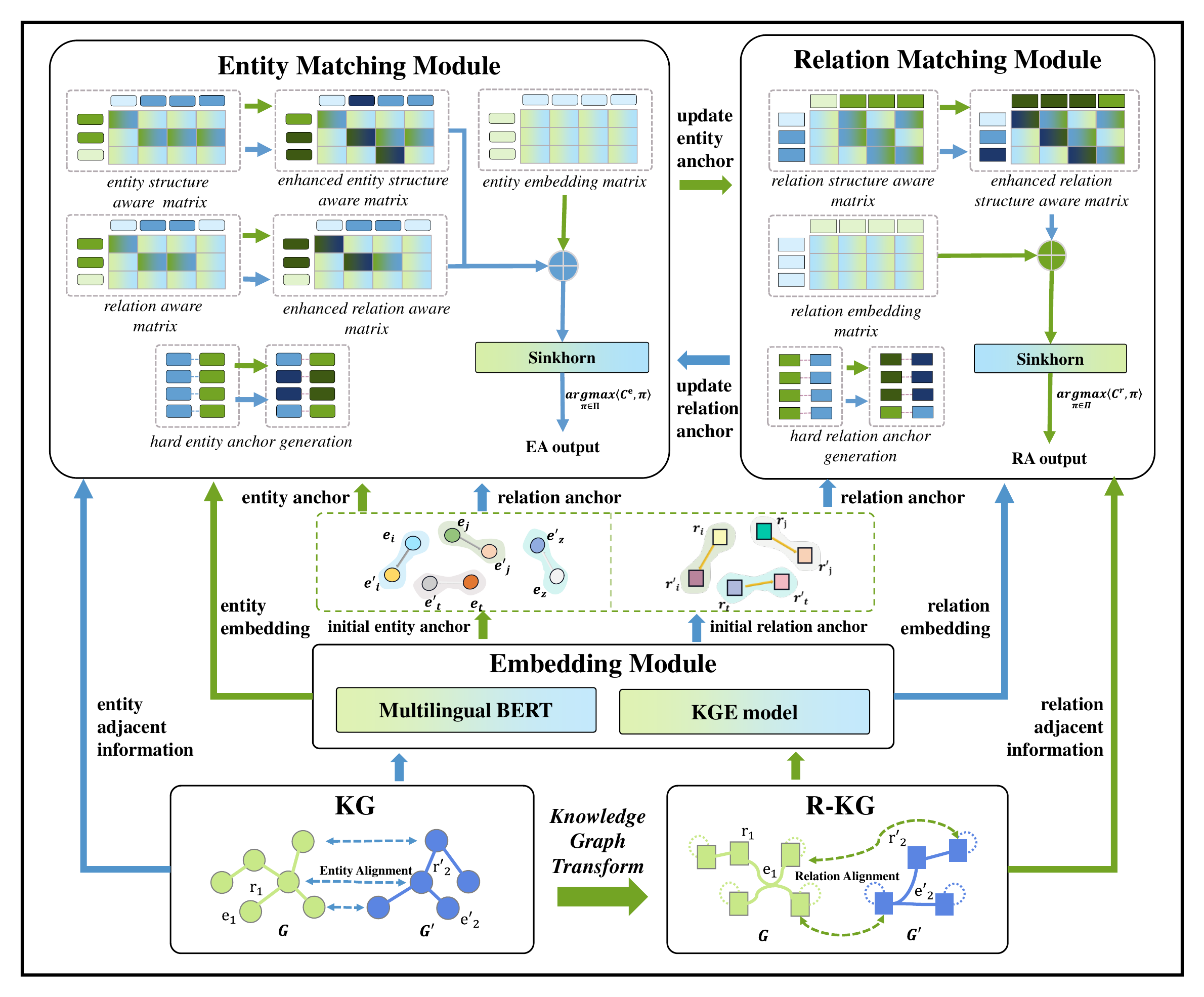}

\end{center}

\caption{The architecture of  EREM.}
\label{fig:EREM}

\end{figure*}

This section provides a detailed exposition of the proposed EREM framework. 
As depicted in Figure \ref{fig:EREM}, EREM comprises three essential components: the hybrid embedding module, the entity matching module, and the relation matching module. 
The hybrid embedding module is responsible for encoding entities and relations into low-dimensional representation spaces and generating the initial sets of entity anchors and relation anchors.
The entity matching module and relation matching module are designed to align entities and relations across KGs.
These two matching modules are trained jointly through an iterative process that alternates between an E-step (EA) and an M-step (RA). 



\subsection{Hybrid Embedding Module}
\label{sec:em}

The hybrid embedding module is designed to generate embeddings for entities and relations derived from both KGs. 
Existing EA models have employed various types of entity encoders, encompassing pure text-based models (e.g., BERT, RoBERTa, and DeBERTa) as well as relation-aware models (e.g., TransD, TransH, TransR, PTransE). 
To accommodate this diversity and enhance generalizability, the hybrid embedding module is constructed to support multiple types of entity and relation encoders. 
For entities, the module integrates both text-based and relation-aware encoders. The embeddings of relations are learned by encoding their titles using multilingual language models.

Formally, The selected embedding module $g$ aims to embed the entity $e$ and relation $r$ into embedding $h_e$ and $h_r$ as follows:
\begin{equation}
\label{eq-init-emb}
    h_e = g(e), h_r = g(r). 
\end{equation} 

Hybrid embedding module is capable of leveraging different encoder types, thereby facilitating the generalization of the proposed training framework.







 

\textbf{Initial Anchor Set Generation.} 
 our work uses embedding module to automatically generate entity and relation anchors.
 The relation and entity anchors will be predicted based on the cosine similarity measure of their embeddings.
 In our approach, our work calculates entity embedding matrix $C^{ent}$ and the relation embedding matrix $C^{rel}$ between two KGs as follows:
\begin{equation} \label{eq4}
\begin{aligned}
    C_{i,j}^{\text {ent }} & =1-\cos (  h_{e_i},  h_{e_j^{\prime}} ), \\
 C_{i,j}^{\text {rel }} & =1-\cos (  h_{r_i},  h_{r_j^{\prime}} ). 
 \end{aligned}
\end{equation}

And, let $C^{ent}_{i,j}$ and $C^{rel}_{i,j}$ < 0.3 as initial entity anchor $(e_i, e'_j)$ and initial relation anchor $(r_i, r'_j)$. To ensure one-to-one align, $C^{ent}_{i,j}$ and $C^{rel}_{i,j}$ must be the minimize in its row.
Then,  embedding module will output entity anchor $\overline{y_e}$ and relation anchor $\overline{y_r}$.


\subsection{Entity Matching Module}
\label{sec:rmm}
This work defines the EA task as an Optimal Transport (OT) problem, wherein the objective is to minimize the global transportation distance.
Given $m$ entities in $G$ and $n$ entities in $G'$, entity cost matrix $C^{e} \in \mathbb{R}^{m \times n}$ is calculated by the sum of entity embedding matrix  $C^{ent}$,  relation-aware matrix $C^{rel}$, and entity structure-aware matrix $C^{stru}$.  
The process of OT alignment is formalized as:
\begin{equation} \label{eq1}
\begin{aligned}
   \min &  \sum_{i=1, j=1}^{n, m} C_{i j}^e \Psi_{i j}^e \\
  C^e &= C^{ent} + C^{stru} + C^{rel}
  \end{aligned}
\end{equation}
\noindent where $\Psi^e$ is the entity transport matrix.  

Given the input of entity and relation anchors, a direct method to enhance entity alignment through relation alignment involves obtaining high-quality entity anchors contingent upon the condition of aligned relations. These high-quality entity anchors can then be utilized to optimize ground truth correspondences, thereby mitigating the noise present in raw entity anchors.

This approach is inspired by a probabilistic scenario: ``If two entities from different knowledge graphs, sharing the same relation, are connected to the same entities, there is a high likelihood that these two entities are identical''. This probabilistic scenario underscores the significance of aligned relations in accurately identifying corresponding entities across different knowledge graphs. 

In detail, given entity anchor $ \overline{y}_e $ and relation anchor $ \overline{y}_r $,  if  $ \{({e_i}, {e_i}^{\prime} ),({e_j}, {e_j}^{\prime})\} \in \overline{y}_e$, $({r_i}, {r_i}^{\prime} )  \in \overline{y}_r $, $({e_i}, {r_i},{e_j}) $  is a triple in $G$ and $ ({e_i}^{\prime} , {r_i}^{\prime}, {e_j}^{\prime} ) $  is another triple in $G^{\prime}$, $ ({e_i}, {e_i}^{\prime} ) $ see as a hard entity anchor $ \widehat{y_e} $  ($ \widehat{y_e} \subset \overline{y_e}$  ).

Given that our alignment objective is to maximize ground truth correspondences, our work formulates our entity alignment objective as the minimization of the negative log-likelihood of entity anchors and relation-aware hard entity anchors. 
So, the optimization function for EA is as follows:

\begin{small}
\begin{equation}
\label{eq-op-em}
    \mathcal{O}^e = -\sum_{(e_i, e_j^{\prime}) \in \overline{y}_e} \log (\Psi_{i j}^e ) - \lambda \cdot \sum_{(e_i, e_j^{\prime}) \in \widehat{y_e}} \log (\Psi_{i j}^e ),
\end{equation}
\end{small}

\noindent where $\lambda \in [0, 1]$ means a  trade-off hyperparameter.


In particular, our work designs a strategy to award  $C^{stru}$ and $C^{rel}$. ($C^{stru} = 1 - S^{stru}$ , $C^{rel} = 1 - S^{rel}$, $S^{stru}$ and $S^{rel}$ are initialized as zero matrix ).  
 The $ S^{stru}$,  $S^{rel}$ are calculated as follows Equation \ref{eq6}.

\begin{equation} \label{eq6}
\begin{aligned}
S_{i,j}^{stru} = \left\{
    		\begin{aligned}
    		S_{i,j}^{stru} + 1, situation I \\ 
    		S_{i,j}^{stru} + \alpha , situation II
    	\end{aligned}
    	\right. \\
S_{i,j}^{rel} = \left\{
    		\begin{aligned}
    		S_{i,j}^{rel} + 1, situation III \\ 
    		S_{i,j}^{rel} + \alpha , situation IV
    	\end{aligned}
    	\right.
\end{aligned}
\end{equation}

\noindent where Situation I is $\{ (e_i, e_j') \in \overline{y_e}, (e_x, e_z') \in \overline{y_e} \}$, Situation II is $\{ (e_i, e_j') \in \widehat{y_e}, (e_x, e_z') \in \widehat{y_e} \}$, Situation III is added condition $ \{ (e_i,r_i,e_x) \in G, (e_j', r_i', e_z') \in G', $ and $ (r_i, r_i') \in \overline{y_r} \} $ on Situation I and Situation IV is added same condition (as Situation III) on Situation II.   $\alpha$ is award value.  $e_x$ and $e_z'$ are the 1-hop entity of $e_i$ and $e_j'$. 
After solving $\Psi^e$, let $(e_i, e'_j)$ as a new anchor to update $\overline{y_e}$ when $\Psi_{i j}^e > \frac{1}{\max(m, n)} - \epsilon$.

In addition, our work will discuss how to apply LLMs in EA and how RA enhances EA in LLM condition in Appendix \ref{sec:appendix-llm}. Referring to the previous works, our work designs the EA procedure as a CoT processing. 
The EA answer responding from LLM will be seen as $\widehat{y_e}$ to enhance EA.  The module will update $\overline{y_e}$ and  $\widehat{y_e}$.

\subsection{Relation Matching Module}
\label{sec:emm}

RA task also be seen as an OT problem in Equation \ref{eq8}. 
Given m relations in $G$ and $n$ relations in $G'$, relation cost matrix $C^{r} \in \mathbb{R}^{m \times n}$ is calculated by the sum of relation embedding matrix  $C^{rel}$,  relation structure aware matrix  $C^{stru\_rel}$ as follows:
\begin{equation} 
\label{eq7}
  C^r = C^{rel} + C^{stru\_rel}
\end{equation}

\begin{equation} \label{eq8}
 \min  \sum_{i=1, j=1}^{n, m} C_{i j}^r \Psi_{i j}^r
\end{equation}

\noindent where $\Psi^r$ is the relation transport matrix. The input of this module contains $\widehat{y_e}$, $\overline{y_e}$, and $\overline{y_r}$, which $\widehat{y_e}$ and $\overline{y_e}$ are getting from embedding matching module. Similar to the E-step, the idea of EA enhancing RA is to fetch hard relation anchors using entity anchors.
In particular, according to $\widehat{y_{e}}$, when $ \{( {e_i}, {e_i}^{\prime} )$, $({e_j}, {e_j}^{\prime}) \} \in \widehat{y_e}$, $( {r_i}, {r_i}^{\prime} )  \in \overline{y_r}$, $ ({e_i}, {r_i} ,{e_j})$ is a triple in $G$, and $ ( {e_i}^{\prime}, {r_i}^{\prime}, {e_j}^{\prime})$ is another triple in $G^{\prime}$,   $ ( {r_i}, {r_i}^{\prime}) $ see as a hard relation anchor $\widehat{y_r}$ ($ \widehat{y_r} \subset \overline{y_r}$  ). Also, the optimization function for RA is as follows: 

\begin{small}
\begin{equation}
\label{eq-op-rm}
    \mathcal{O}^r = -\sum_{(r_i, r_j^{\prime}) \in \overline{y}_r} \log (\Psi_{i j}^r ) - \lambda \cdot \sum_{(r_i, r_j^{\prime}) \in \widehat{y_r}} \log (\Psi_{i j}^r ),
\end{equation}
\end{small}

To achieve it, $S^{stru\_rel} $ is calculated as follows:

\begin{equation} \label{eq10}
S_{i,j}^{stru\_rel} = \left\{
    		\begin{aligned}
    		S_{i,j}^{stru\_rel} + 1, situation V \\ 
    		S_{i,j}^{stru\_rel} + \alpha , situation VI
    	\end{aligned}
    	\right. 
\end{equation}

\noindent where Situation V is $\{ (r_i, r_j') \in \overline{y_r},  (r_x, r_z') \in \overline{y_r} \}$ , Situation VI is $\{ (r_i, r_j') \in \widehat{y_r} , (r_x, r_z') \in \widehat{y_r}   \}$,  $r_x$ and $r_z'$ are the 1-hop neighborhood entity of $r_i$ and $r_j'$.  relation structure  matrix $C^{stru\_rel}$($C^{stru\_rel} = 1 - S^{stru\_rel}$ ).  After solving $\Psi^r$, let $(r_i, r'_j)$ as a new anchor to update $\overline{y_r}$ when $\Psi_{i j}^r > \frac{1}{\max(m, n)} - \epsilon$.
However, it will lead to a high complexity time to compute $C^{stru\_rel}$ with the algorithm complexity of $O(n^2)$.  
 To facilitate the catch of global relation structure information,  \textit{Knowledge Graph Transformation} (KGT) is proposed to transform the original KG ( entity1, relation, entity2 ) into R-KG ( relation1, entity, relation2 ), and this process can be shown in Figure \ref{fig:EREM} (In subfigure of KG and R-KG). 
 R-KG can efficiently and consistently retrieve comprehensive global context information for relations. 
In addition, the complexity of obtaining the adjacency of relations can be reduced to $O(n)$. 
Finally, our work will discuss how to apply LLMs in RA and how EA enhances RA in LLM condition in Appendix \ref{sec:appendix-llm}. 
Similar to the EA, our work designs the RA procedure as a CoT processing. 
The RA answer responding from LLM will be seen as $\widehat{y_r}$ to enhance RA.  The module will update $\overline{y_r}$
and  $\widehat{y_r}$.

\renewcommand{\algorithmiccomment}[1]{/* #1 */}

\begin{algorithm}[t]
    \begin{small}
  \caption{The Procedure of our EREM}
   \label{alg:algo}
\begin{algorithmic}[1]
   \STATE {\bfseries Input:}  two KGs $G$ and $G^{\prime}$
    \STATE {\bfseries Parameter:}  iteration number $T$,  entity anchors $\overline{y}_e$, relation anchors $\overline{y}_r$, hard entity anchors $\widehat{y}_e$, hard relation anchors $\widehat{y}_e$
   
    \STATE \begin{varwidth}[t]{\linewidth}
             Encode entities and relations using Eq (\ref{eq-init-emb});  \par
        \end{varwidth} 
\STATE \begin{varwidth}[t]{\linewidth}
            Generate the entity anchors $\overline{y}_e$ and relation anchors $\overline{y}_r$; \par
        \end{varwidth} 
        \STATE \begin{varwidth}[t]{\linewidth}
            Initialize Entity Matching Module and Relation Matching Module; \par
        \end{varwidth} 
   \WHILE{$k = 1, ..., T$}
        \STATE \begin{varwidth}[t]{\linewidth}
             Generate $\widehat{y}_e$ based on  $\overline{y}_r$;  \par
        \end{varwidth} 

        \STATE \begin{varwidth}[t]{\linewidth}
             Train Entity Matching Module optimizing by minimizing Eq (\ref{eq-op-em});  \par
        \end{varwidth} 

        \STATE \begin{varwidth}[t]{\linewidth}
             Update entity anchors;  \par
        \end{varwidth} 

        \STATE \begin{varwidth}[t]{\linewidth}
             Test EA;  \par
        \end{varwidth} 

        \STATE \begin{varwidth}[t]{\linewidth}
             Generate $\widehat{y}_r$ based on $\widehat{y}_e$ ;  \par
        \end{varwidth} 

        \STATE \begin{varwidth}[t]{\linewidth}
             Train Relation Matching Module optimizing by minimizing Eq (\ref{eq-op-rm});  \par
        \end{varwidth} 

        \STATE \begin{varwidth}[t]{\linewidth}
             Update relation anchors;  \par
        \end{varwidth} 
        
        \STATE \begin{varwidth}[t]{\linewidth}
             Test RA;  \par
        \end{varwidth} 
   \ENDWHILE
    \STATE {\bfseries Return:} EA output and RA output
\end{algorithmic}
\end{small}
\end{algorithm}

\subsection{ EREM: Joint of Entity Matching and Relation Matching  with E-step and M-step}
\label{sec:EREM}
The joint training of Entity Matching and Relation Matching utilizes a variational EM procedure, optimized through the alternation between an E-step and an M-step.
In the E-step, EA is optimized utilizing the relation anchor learned by RA, with the objective of maximizing entity anchor correspondences.
In the M-step, 
RA is optimized by leveraging the entity anchors predicted by EA, with the objective of maximizing the correspondences between relation anchors.
Ultimately, our work demonstrates the final objective by incorporating the EA and RA through summation:

\begin{equation}
\label{eq11}
    \mathcal{O}^{final} = \mathcal{O}^e + \mathcal{O}^r
\end{equation}

In the subsequent iterations ($t = 1, ... T$), the Entity Matching and the Relation Matching modules are alternately performed. 
Through each iterator, our work uses $\overline{y_{r}}$ to create  $\widehat{y_e}$  to boost EA in Entity Matching module and  $\widehat{y_e}$  additionally use to create $\widehat{y_r}$ to boost RA in Relation Matching module.
Algorithm \ref{alg:algo} shows the whole process of the EREM framework.

\section{Experiments}

\subsection{Experimental Settings}

\begin{table}[htbp]

\caption{Dataset statistics.} 
\label{datasets}
\resizebox{\hsize}{!}
{
\begin{tabular}{llllll}
\toprule
\multirow{2}{*}{Dataset}     & \multirow{2}{*}{Lang.} & \multicolumn{2}{l}{Entity}       & \multicolumn{2}{l}{Relation}  \\ \cline{3-6} 
                             &                        & Num.   & \#Align.                & Num.  & \#Align.              \\  \midrule
\multirow{2}{*}{DBP15K\_ZH-EN} & ZH                     & 19,388 & \multirow{2}{*}{15000}  & 1,701 & \multirow{2}{*}{997} \\
                             & EN                     & 19,572 &                         & 1,323 &                       \\  \hline
\multirow{2}{*}{DBP15K\_JA-EN} & JA                     & 19,814 & \multirow{2}{*}{15000}  & 1,299 & \multirow{2}{*}{684} \\
                             & EN                     & 19,780 &                         & 1,153 &                       \\ \hline
\multirow{2}{*}{DBP15K\_FR-EN} & FR                     & 19,661 & \multirow{2}{*}{15000}  & 903   & \multirow{2}{*}{274}  \\
                             & EN                     & 19,993 &                         & 1,208 &                       \\ 
                          
                             \bottomrule

\end{tabular}
}

\end{table}

\noindent \textbf{Dataset}
The DBP15K\_{ZH-EN} dataset comprises 997 aligned relation pairs, while the DBP15K\_{JA-EN} and DBP15K\_{FR-EN} datasets contain 684 and 274 aligned relation pairs, respectively. Detailed information about these datasets is provided in Table \ref{datasets}. 
To validate both the EA and RA tasks, our work manually labeled the data for relation alignment within the DBP15K dataset. 


 \begin{table*}[htbp]
\caption{The overall EA results of the EREM models in comparison with the baselines. (G$\uparrow$ denotes the improvements of EREM over the baseline model on EA task.)}
\label{tab:main_ea}
\resizebox{\textwidth}{!}
{
\begin{tabular}{cccccccccccc}
\hline
\multirow{2}{*}{Model/G$\uparrow$  } & \multicolumn{3}{c}{DBP15K-ZH\_EN} & \multicolumn{3}{c}{DBP15K-JA\_EN} & \multicolumn{3}{c}{DBP15K-FR\_EN} & \multicolumn{2}{c}{AVG}   \\ \cline{2-12} 
                         & Hits@1    & Hits@10    & MRR      & Hits@1    & Hits@10    & MRR      & Hits@1    & Hits@10    & MRR      & Hits@1      & MRR         \\ \hline
KECG                     & 0.467     & 0.815      & 0.586    & 0.485     & 0.843      & 0.605    & 0.476     & 0.849      & 0.601    & 0.476       & 0.597       \\
KECG-EREM               & 0.822         & 0.946   & 0.867 & 0.800         & 0.941   & 0.849 & 0.832         & 0.959   & 0.878 & 0.818       & 0.865      \\
G$\uparrow$                   & \colorbox{tealcolor}{0.355}        & \colorbox{tealcolor}{0.131}   & \colorbox{tealcolor}{0.281} & \colorbox{tealcolor}{0.315}         & \colorbox{tealcolor}{0.098}   & \colorbox{tealcolor}{0.244} & \colorbox{tealcolor}{0.356}         & \colorbox{tealcolor}{0.110}   & \colorbox{tealcolor}{0.277} & \colorbox{tealcolor}{0.342}       & \colorbox{tealcolor}{0.267}       \\ \hline
GCNAlign                 & 0.410   &	0.756   &	0.527   &	0.442   &	0.810   &	0.566   &	0.430   &	0.813   &	0.557    &	0.427   &	0.550
      \\
GCNAlign-EREM            & 0.751         & 0.925   & 0.811 & 0.740         & 0.917   & 0.802 & 0.799         & 0.951   & 0.853 & 0.763       & 0.822      \\
G$\uparrow$                      &   \colorbox{tealcolor}{0.341}  &  	\colorbox{tealcolor}{0.169}  &  	\colorbox{tealcolor}{0.284}  &  	\colorbox{tealcolor}{0.298}  &  	\colorbox{tealcolor}{0.107} &  	\colorbox{tealcolor}{0.236} &  	\colorbox{tealcolor}{0.369} &  	\colorbox{tealcolor}{0.138} &  	\colorbox{tealcolor}{0.296} &  	\colorbox{tealcolor}{0.336} &  	\colorbox{tealcolor}{0.272}
   \\ \hline
RotatE                   & 0.423     & 0.754      & 0.534    & 0.448     & 0.785      & 0.561    & 0.439     & 0.800      & 0.560    & 0.437       & 0.552       \\
RotatE-EREM              & 0.678     & 0.872      & 0.745    & 0.649     & 0.869      & 0.724    & 0.745     & 0.929      & 0.810    & 0.691       & 0.760       \\
G$\uparrow$                          & \colorbox{tealcolor}{0.255}     & \colorbox{tealcolor}{0.118}      & \colorbox{tealcolor}{0.211}    & \colorbox{tealcolor}{0.201}     & \colorbox{tealcolor}{0.084}     & \colorbox{tealcolor}{0.163}    & \colorbox{tealcolor}{0.306}     & \colorbox{tealcolor}{0.129}      & \colorbox{tealcolor}{0.250}    & \colorbox{tealcolor}{0.254}       & \colorbox{tealcolor}{0.208}       \\ \hline
BootEA                   & 0.490     & 0.793      & 0.593    & 0.499     & 0.813      & 0.605    & 0.515     & 0.838      & 0.623    & 0.501       & 0.607       \\
BootEA-EREM              & 0.770     & 0.912      & 0.820    & 0.767     & 0.915      & 0.820    & 0.805     & 0.948      & 0.857    & 0.781       & 0.832       \\
G$\uparrow$                          & \colorbox{tealcolor}{0.280}     & \colorbox{tealcolor}{0.119}      & \colorbox{tealcolor}{0.227}    & \colorbox{tealcolor}{0.268}     & \colorbox{tealcolor}{0.102}      & \colorbox{tealcolor}{0.215}    & \colorbox{tealcolor}{0.290}     & \colorbox{tealcolor}{0.110}      & \colorbox{tealcolor}{0.234}    & \colorbox{tealcolor}{0.279}       & \colorbox{tealcolor}{0.225}       \\ \hline
BERT-INT                 & 0.968     & 0.990      & 0.977    & 0.964     & 0.991      & 0.975    & 0.992     & 0.998      & 0.995    & 0.975       & 0.982       \\
BERTINT-EREM             & \textbf{0.995}     & \textbf{1.000}      & \textbf{0.997}    & \textbf{0.993 }    & \textbf{1.000}      & \textbf{0.996 }   & 0.996     & 1.000      & 0.998    & \textbf{0.995}       & \textbf{0.997}       \\
G$\uparrow$                          & \colorbox{tealcolor}{0.027}     & \colorbox{tealcolor}{0.010}      & \colorbox{tealcolor}{0.020}    & \colorbox{tealcolor}{0.029}     & \colorbox{tealcolor}{0.009}      & \colorbox{tealcolor}{0.021}    & \colorbox{tealcolor}{0.004}     & \colorbox{tealcolor}{0.002}      & \colorbox{tealcolor}{0.003}    & \colorbox{tealcolor}{0.020}       & \colorbox{tealcolor}{0.015}       \\ \hline
FGWEA                    & 0.976     & 0.994      & 0.983    & 0.978     & 0.992      & 0.988    & 0.997     & 0.999      & 0.998    & 0.984       & 0.990       \\
FGWEA-EREM               & 0.980     & 0.995      & 0.985    & 0.979     & 0.993      & 0.989    & \textbf{0.999}     & \textbf{1.000}      & \textbf{0.999}    & 0.986       & 0.991       \\
G$\uparrow$                          & \colorbox{tealcolor}{0.004}     & \colorbox{tealcolor}{0.001}      & \colorbox{tealcolor}{0.002}    & \colorbox{tealcolor}{0.001}     & \colorbox{tealcolor}{0.001}      & \colorbox{tealcolor}{0.001}    & \colorbox{tealcolor}{0.002}     & \colorbox{tealcolor}{0.001}      & \colorbox{tealcolor}{0.001}    & \colorbox{tealcolor}{0.002} & \colorbox{tealcolor}{0.001} \\ \hline

ChatGLM6B                    & 0.882 & 	0.953 & 	0.909 & 	0.903 &  	0.960  & 	0.924 & 	0.987 & 	0.998 & 	0.992 & 	0.928 & 	0.942 
      \\
ChatGLM6B-EREM               & 0.972 & 	0.993 & 	0.98 & 	0.978  & 	0.992 & 	0.983 &  	0.998 & 	1.000 & 	0.999 & 	0.983 & 	0.987 
       \\
G$\uparrow$                          & \colorbox{tealcolor}{0.09} & 	\colorbox{tealcolor}{0.04} & 	\colorbox{tealcolor}{0.071} & 	\colorbox{tealcolor}{0.075}	& \colorbox{tealcolor}{0.032} & 	\colorbox{tealcolor}{0.059} & 	\colorbox{tealcolor}{0.011} & 	\colorbox{tealcolor}{0.002} & 	\colorbox{tealcolor}{0.007} & 	\colorbox{tealcolor}{0.056} & 	\colorbox{tealcolor}{0.046}
 \\ \hline

\end{tabular}
}
\end{table*}

\noindent \textbf{Evaluation Metrics}
Following previous works \cite{10.1145/3459637.3482338}, the evaluation metrics for EA and RA include Hits@k ($k = 1, 10$) and Mean Reciprocal Rank (MRR). 


\noindent \textbf{Baselines}
our work employs seven competitive KG alignment methods as baselines, including both foundational KGE-based techniques and recent advanced methods.
KECG \cite{li2019semi}, GCNAlign \cite{wang2018cross}, RotatE \cite{sun2018rotate}, and BootEA \cite{sun2018bootstrapping} are KEG-based methods. 
BERT-INT \cite{ijcai2020p439} is the SOTA bert-based method.
 FGWEA \cite{tang2023fused} follows ``embedding module and entity matching module''  strategy which is the SOTA unsupervised method.
Incorporating LLMs into our work, our work utilizes ChatGLM-6b to construct a foundational model as detailed in Appendix \ref{sec:appendix-llm}. 

\noindent \textbf{Implementation Details.} Our work implements our framework using Pytorch. For our methods, the common hyper-parameters are listed as follows: The embedding Module can be LaBSE, KECG, BootEA, RotatE, and GCN-Align, and its embedding dimensions are 768, 128, 75, 200, and 100. our work updates 8 iterations ($T=8$) for EA  OT alignment and  RA OT alignment is solved by Sinkhorn algorithm (entropic regularization weight is set to 0.1). $ \epsilon $ is set to 1e-5.
$\lambda$ is set to 1. The award $\alpha$ is set to 2. When using LaBSE as an embedding module, the name and attribute information of entity and relation are encoded into text semantic vectors. For KECG, BootEA, RotatE, and GCN-Align, 30\% of the supervised labeled entity information is removed to construct the test set to prevent data leakage during the validation stage.  In our experiments, our work don't use a translated version of DBP15K. 
Our experiments are conducted on a workstation with a GeForce GTX 3090 GPU.

\subsection{Main Results}

\begin{table*}[h]
\caption{Ablation study on the different settings of EREM}
\label{tab:ab_study}
\resizebox{\hsize}{!}
{
\begin{tabular}{llllllllll}
\hline
\multirow{2}{*}{Model} & \multicolumn{3}{l}{DBP15K-ZH\_EN} & \multicolumn{3}{l}{DBP15K-JA\_EN} & \multicolumn{3}{l}{DBP15K-FR\_EN} \\ \cline{2-10} 
                       & Hits@1    & Hits@10    & MRR      & Hits@1    & Hits@10    & MRR      & Hits@1    & Hits@10    & MRR      \\ \hline
GCNAlign-EREM          & 0.751     & 0.925      & 0.811    & 0.740     & 0.917      & 0.802    & 0.799     & 0.951      & 0.853    \\
GCNAlign-EREM(-E)      & 0.709     & 0.913      & 0.780    & 0.719     & 0.912      & 0.785    & 0.767     & 0.947      & 0.831    \\
GCNAlign-EREM(-M)      & 0.683     & 0.908      & 0.759    & 0.706     & 0.910      & 0.776    & 0.731     & 0.940      & 0.803    \\ \hline
\end{tabular}
}
\end{table*}

\textbf{Performance Analysis on Entity Alignment.} 
The results of all models on the entity alignment task are shown in Table \ref{tab:main_ea}.  
From the result, our work has the following findings. 
Compared with the strongest baseline FGWEA on all three datasets, our EREM achieves 0.1\% to 0.4\% relative improvements on Hits@1, and 0.1\% relative improvements on Hits@10.
For BERT-INT, EREM achieves 0.4\% to 2.9\% relative improvements on Hits@1, and 0.2\% to 1.0\% relative improvements on Hits@10.
KGE-based models can be seen as embedding model, and use EREM to optimize them. 
Through EREM, KECG can improve the average Hits@1 score by 34.2\%, GCNAlign can improve the average  Hits@1 score by 33.6\%, RotatE can improve the average  Hits@1 score by 25.4\%, and BootEA can improve the average  Hits@1 score by 27.9\%. ChatGLM6b can improve the average Hits@1 score by 5.6\%.
These improvements demonstrate the superiority of our proposal and the effectiveness of joint training EA and RA using EM optimization.


\begin{table*}[htbp]
\caption{The overall RA results of the EREM models in comparison with the baselines (G$\uparrow$ denotes the improvements of EREM over the baseline model on RA task.)}
\label{tab:main_ra}
\resizebox{\textwidth}{!}
{
\begin{tabular}{cccccccccccc}
\hline
\multirow{2}{*}{Model/G$\uparrow$} & \multicolumn{3}{l}{DBP15K-ZH\_EN} & \multicolumn{3}{l}{DBP15K-JA\_EN} & \multicolumn{3}{l}{DBP15K-FR\_EN} & \multicolumn{2}{l}{AVG} \\ \cline{2-12} 
                         & Hits@1    & Hits@10    & MRR      & Hits@1    & Hits@10    & MRR      & Hits@1    & Hits@10    & MRR      & Hits@1      & MRR       \\ \hline
KECG                     & 0.517     & 0.606      & 0.551    & 0.238     & 0.281      & 0.260    & 0.234     & 0.299      & 0.263    & 0.330       & 0.358     \\
KECG-EREM                & 0.649     & 0.714      & 0.676    & 0.409     & 0.475      & 0.440    & 0.336     & 0.453      & 0.411    & 0.465       & 0.509     \\
G$\uparrow$                        & \colorbox{tealcolor}{0.132}     & \colorbox{tealcolor}{0.108}      & \colorbox{tealcolor}{0.125}    & \colorbox{tealcolor}{0.171}     & \colorbox{tealcolor}{0.194}      & \colorbox{tealcolor}{ 0.180}    & \colorbox{tealcolor}{ 0.102}     & \colorbox{tealcolor}{ 0.154}      & \colorbox{tealcolor}{ 0.148}    & \colorbox{tealcolor}{ 0.135}       & \colorbox{tealcolor}{ 0.151}     \\ \hline
GCNAlign                 & 0.339     & 0.383      & 0.360    & 0.333     & 0.409      & 0.369    & 0.215     & 0.321      & 0.261    & 0.296       & 0.330     \\
GCNAlign-EREM            & 0.590     & 0.635      & 0.612    & 0.493     & 0.569      & 0.527    & 0.409     & 0.515      & 0.450    & 0.497       & 0.530     \\
G$\uparrow$                        & \colorbox{tealcolor}{ 0.251}     & \colorbox{tealcolor}{ 0.252}      & \colorbox{tealcolor}{ 0.252}    & \colorbox{tealcolor}{ 0.160}     & \colorbox{tealcolor}{ 0.160}      & \colorbox{tealcolor}{ 0.158}    & \colorbox{tealcolor}{ 0.194}     & \colorbox{tealcolor}{ 0.194}      & \colorbox{tealcolor}{ 0.189}    & \colorbox{tealcolor}{ 0.202}       & \colorbox{tealcolor}{ 0.200}     \\ \hline
RotatE                   & 0.067     & 0.107      & 0.085    & 0.080     & 0.121      & 0.098    & 0.091     & 0.135      & 0.109    & 0.079       & 0.097     \\
RotatE-EREM              & 0.158     & 0.216      & 0.184    & 0.145     & 0.145      & 0.170    & 0.175     & 0.259      & 0.209    & 0.159       & 0.188     \\
G$\uparrow$                        & \colorbox{tealcolor}{ 0.091}     & \colorbox{tealcolor}{ 0.109}      & \colorbox{tealcolor}{ 0.099}    & \colorbox{tealcolor}{ 0.065}     & \colorbox{tealcolor}{ 0.024 }      &  \colorbox{tealcolor}{ 0.072 }   &  \colorbox{tealcolor}{ 0.084 }     & \colorbox{tealcolor}{ 0.124 }      & \colorbox{tealcolor}{ 0.100 }   & \colorbox{tealcolor}{ 0.080}       & \colorbox{tealcolor}{ 0.090}     \\ \hline
BootEA                   & 0.545     & 0.636      & 0.579    & 0.421     & 0.569      & 0.477    & 0.361     & 0.504      & 0.415    & 0.442       & 0.490     \\
BootEA-EREM              & 0.567     & 0.659      & 0.600    & 0.474     & 0.626      & 0.530    & 0.401     & 0.544      & 0.451    & 0.481       & 0.527     \\
G$\uparrow$                        & \colorbox{tealcolor}{ 0.022 }    & \colorbox{tealcolor}{  0.023 }     & \colorbox{tealcolor}{ 0.021}    & \colorbox{tealcolor}{ 0.053 }     &  \colorbox{tealcolor}{ 0.057 }      &  \colorbox{tealcolor}{ 0.053 }   & \colorbox{tealcolor}{ 0.040 }    & \colorbox{tealcolor}{ 0.040 }     & \colorbox{tealcolor}{ 0.036 }    & \colorbox{tealcolor}{ 0.038}       & \colorbox{tealcolor}{ 0.037 }     \\ \hline
BERT-INT              &   0.909 &	0.949 &	0.927 &	0.849 &	0.934 	& 0.884 &	0.682 &	0.825 &	0.735 &	0.815 &	0.849 

      \\
BERTINT-EREM           &    0.911   &	0.949   &	0.927   &	0.849   &	0.934   &	0.884   &	0.686 	  & 0.821   &	0.737   &	0.815 &	0.849

       \\
G$\uparrow$                    &      \colorbox{tealcolor}{ 0.002 }	 & 0.000  &	0.000  &	0.000  &	0.000  &	0.000 	& \colorbox{tealcolor}{ 0.004 } &	\colorbox{lowred}{-0.04} & \colorbox{tealcolor}{ 	0.002}  &	\colorbox{tealcolor}{ 0.002 } &	 \colorbox{tealcolor}{ 0.001 } 
 
     \\ \hline
FGWEA                    & 0.91  & 	0.948  & 	0.927  & 	0.858  & 	0.936  & 	0.89  & 	0.679   & 	0.818  &  	0.736 	  & 0.816  & 	0.851
      \\
FGWEA-EREM               & 0.911  &	0.952  &	0.928  &	0.86  &	0.931  &	0.888  &	0.679   &	0.818  & 	0.736 	 & 0.817   & 0.851
       \\
G$\uparrow$                          & \colorbox{tealcolor}{ 0.001} & 	\colorbox{tealcolor}{ 0.004} & 	\colorbox{tealcolor}{ 0.001} & 	\colorbox{tealcolor}{ 0.002} & \colorbox{lowred}{ 	-0.005} & 	\colorbox{lowred}{-0.002} & 	0 & 	0 & 	0	 & \colorbox{tealcolor}{ 0.001} & 	0
 \\ \hline
 ChatGLM6b                    & 0.923	 & 0.969 &	0.93 &	0.863 &	0.939 &	0.898 &	0.679  &	0.818  &	0.736 & 	0.822 &	0.855

      \\
ChatGLM6b-EREM               & \textbf{0.925}	 & \textbf{0.971} & 	\textbf{0.932} & 	\textbf{0.866} & 	\textbf{0.944} & 	\textbf{0.901}	 & \textbf{0.679} 	 & \textbf{0.818}  & 	\textbf{0.736}  & 	\textbf{0.823}  & 	\textbf{0.856}

       \\
G$\uparrow$                          & \colorbox{tealcolor}{ 0.002} &	\colorbox{tealcolor}{ 0.002} &	\colorbox{tealcolor}{ 0.002 } &	\colorbox{tealcolor}{ 0.003} &	\colorbox{tealcolor}{ 0.005} &	\colorbox{tealcolor}{ 0.003} &	0 &	0 &	0 & 	\colorbox{tealcolor}{ 0.002} &	\colorbox{tealcolor}{ 0.002}

 \\ \hline
\end{tabular}
}
\end{table*}

\begin{figure*}[h]
\begin{minipage}[t]{0.33\linewidth}
        \centering
        \includegraphics[scale=0.16]{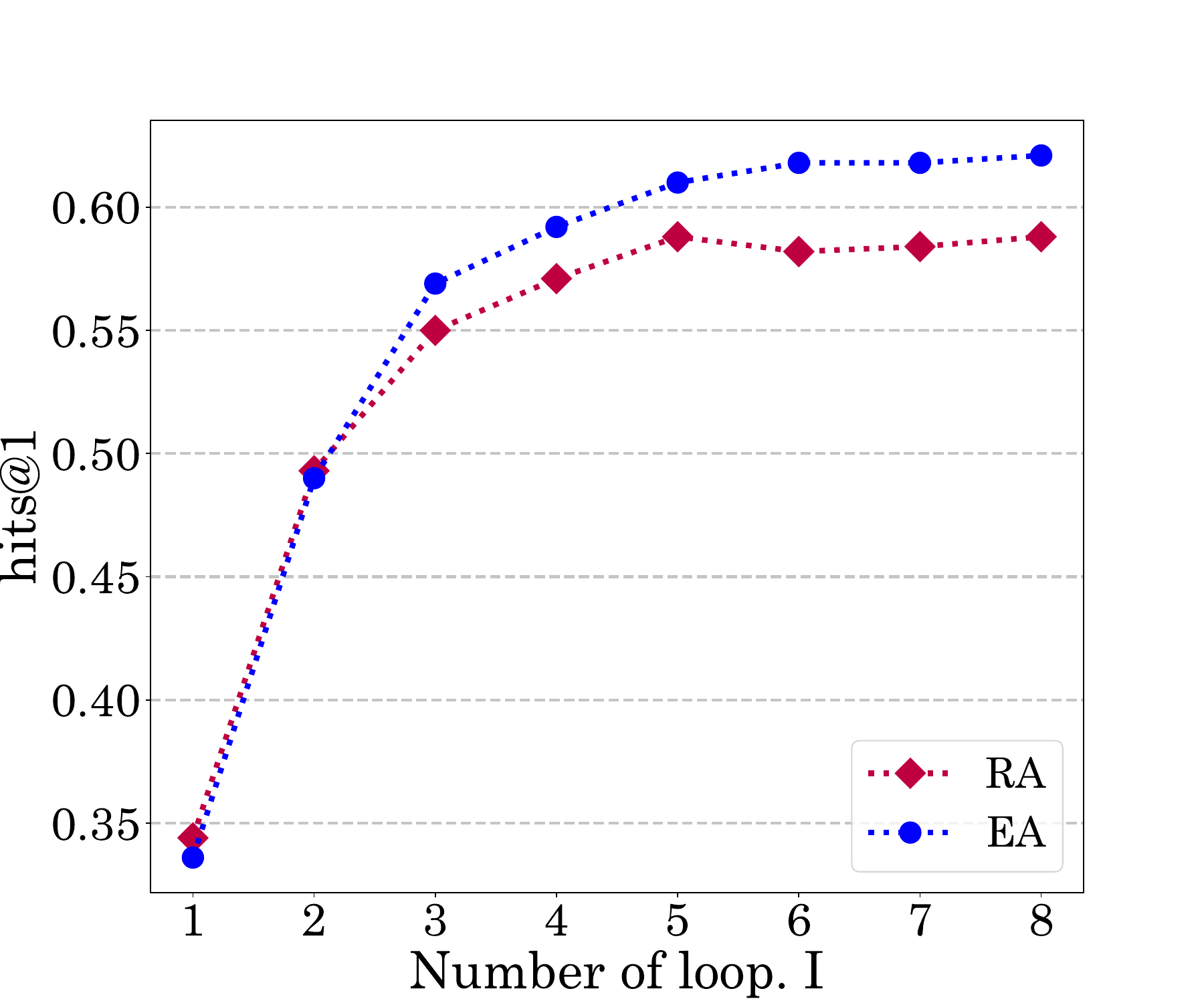}
        \label{fig:ea_ra_1}
    \end{minipage}
        \hfill  
    \begin{minipage}[t]{0.33\linewidth}
        \centering
        \includegraphics[scale=0.16]{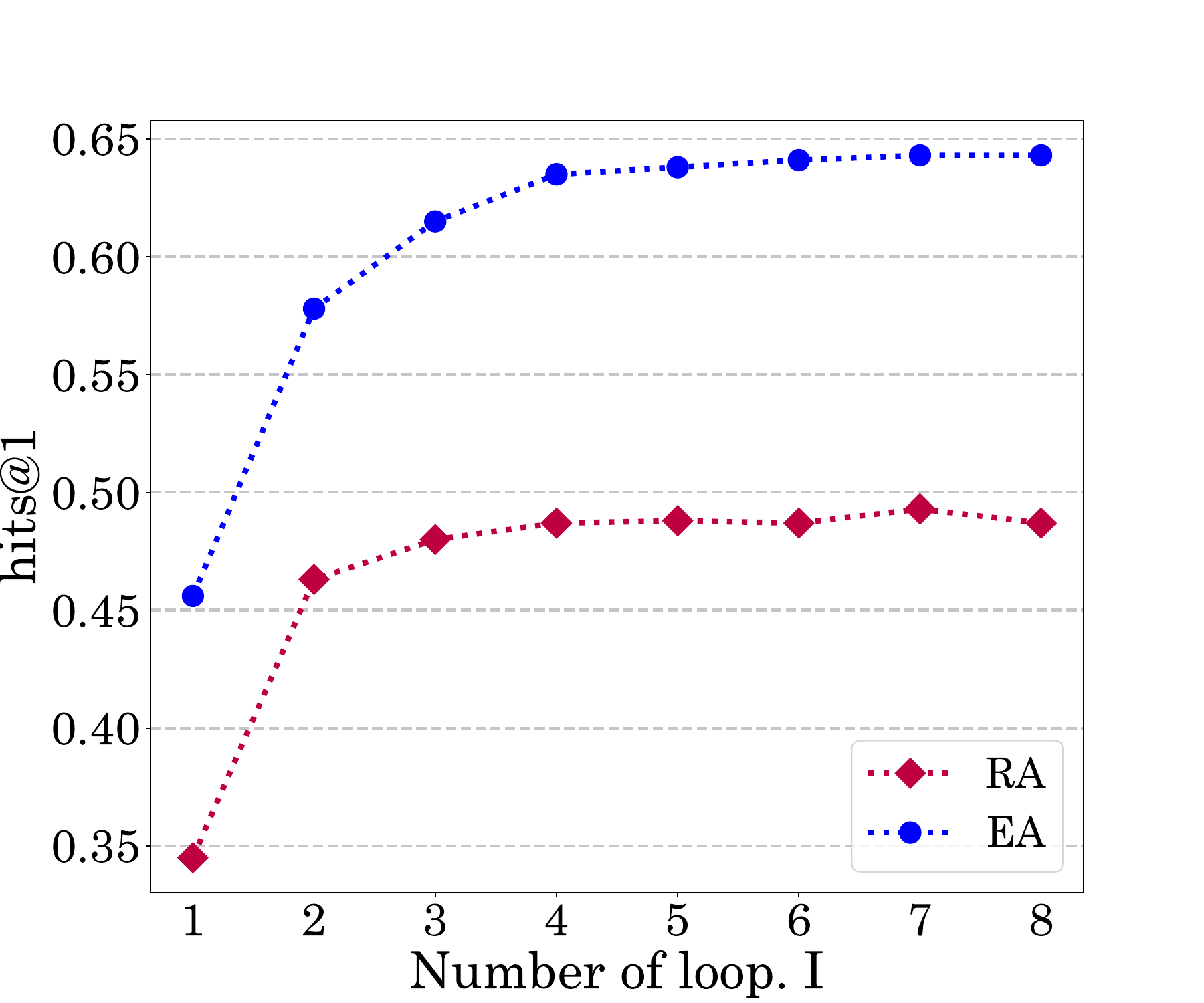}
        \label{fig:ea_ra_2}
    \end{minipage}%
    \hfill  
    \begin{minipage}[t]{0.33\linewidth}
        \centering
        \includegraphics[scale=0.16]{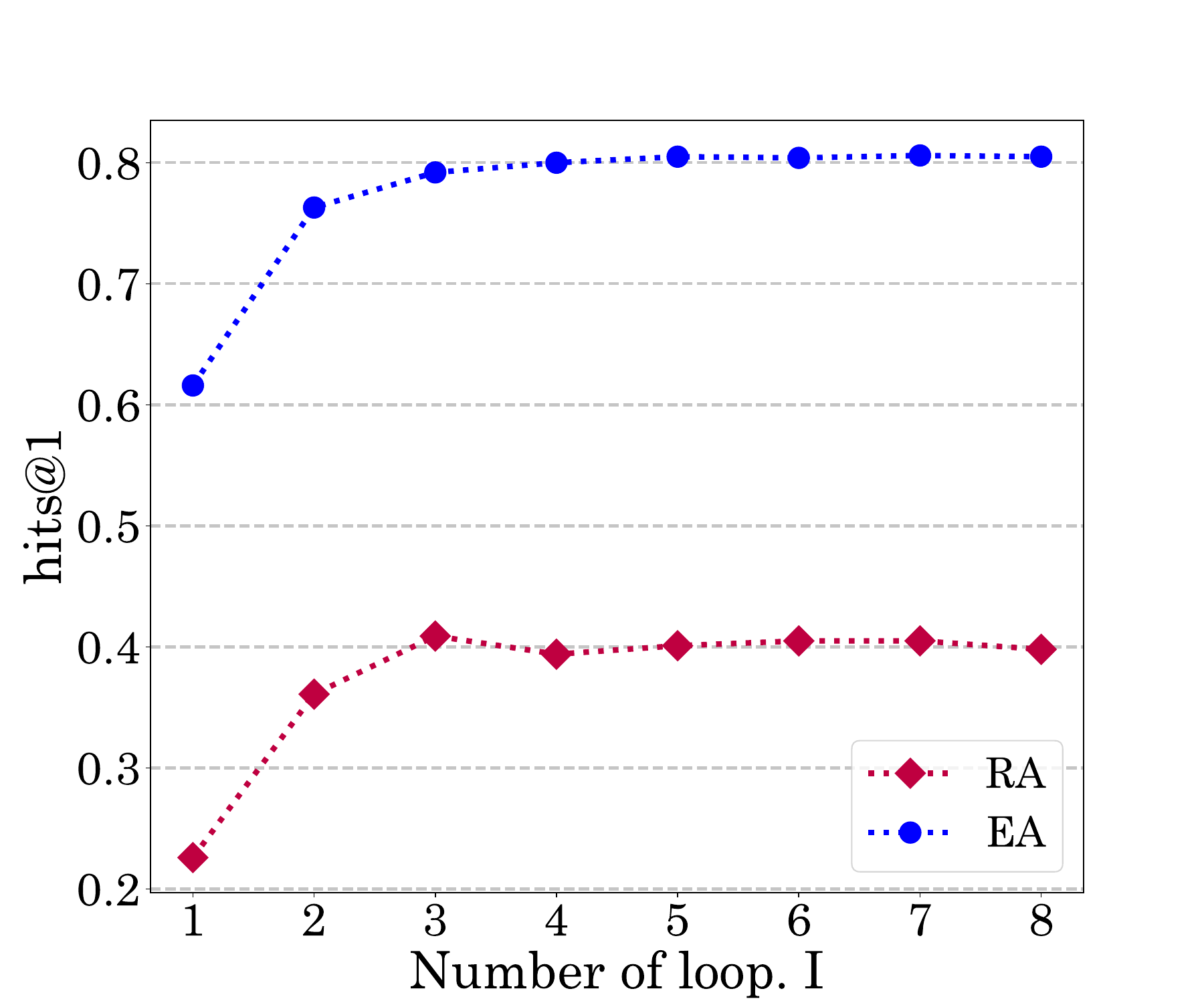}
        \label{fig:ea_ra_3}
    \end{minipage}
   
    \caption{ 
The three sets of figures illustrate the mutual promotion of entity alignment and relation alignment during the training iteration process(from Iter(I)=1 to 8), resulting in a collective improvement in training accuracy (e.g. the embedding module are GCN-Align). 
}
 \label{fig:ea_ra}
\end{figure*}

\noindent \textbf{Performance Analysis on Entity Alignment}.   
The EREM model, proposed for entity alignment, extends its applicability to RA as well. 
In Table \ref{tab:main_ra}, our work presents the comparative results across three datasets for RA.
Applying EREM for optimization purposes yields significant improvements in RA. Specifically, KECG shows an enhancement from 10.2\% to 17.1\% in  Hits@1, GCN-Align experiences an improvement from 16.0\% to 25.1\% in  Hits@1, RotatE experiences an improvement from 6.5\% to 9.1\% in  Hits@1, and BootEA sees an increase from 2.2\% to 5.3\% in  Hits@1. 
For FGWEA and BERT-INT, EREM can only get a few (0.1\% and 0.2\%) improvements in the average of  Hits@1.  ChatGLM6b can improve the average Hits@1 score by 0.2\%. 
A few declines on H@10 and MRR may be caused by the relation evaluation sets, which contain some ambiguous relations making model hard to recognize.
These results underscore the effectiveness of EREM in enhancing the performance of these models in RA.
These results further support our assumption that the alignment of entities and relations within our model can mutually reinforce each other.

\subsection{Ablation Study}
To evaluate the effectiveness of our designed modules, our work design two ablation models as follows: 
GCNAlign-EREM (-E) denotes GCNAlign-EREM without considering the impact of relation alignment on entity alignment in the E-step, 
and GCNAlign-EREM (-M) denotes the GCNAlign-EREM without considering the impact of entity alignment on relation alignment in the M-step. 
Table \ref{tab:ab_study} demonstrates the experimental results.   
The full EREM model outperforms these two variants, validating the effectiveness of the proposed progressive EM optimization algorithm. 
The results indicate that excluding the E-step, thereby not using RA to enhance EA, results in a performance decrease of 2.1\% to 4.2\%. 
Similarly, excluding the M-step leads to a performance decline of 3.4\% to 6.8\%. 
Therefore, the iterative training of the entity matching module and relation matching module using EREM is effective for both EA and RA.

\subsection{Hyper-parameter Sensitivity Analysis}

To gain a comprehensive understanding of the extent to which RA can enhance EA,  Figure \ref{fig:ea_ra} presents the results of both EA and RA throughout the iterative process of our experiments.  

One can clearly see that with the increase of the training iterations, the performance of both EA and RA increases. 
These results demonstrate the mutual enhancement of EA and RA within the proposed framework.
Figure  \ref{fig:hard_anchor} (Appendix \ref{sec:appendix}) illustrates the increase in the number and accuracy of hard entity anchors during the training process, highlighting their superior performance compared to normal entity anchors.
To compare the impact of Expectation-Maximization (EM) enhancement on the entity cost matrix, our work uses heatmaps to represent the changes before and after applying EM in Figure \ref{fig:hot_1} and \ref{fig:hot_2} (Appendix \ref{sec:appendix}). 
These heatmaps indicate that the EREM places more emphasis on hard entity anchors as training converges, underscoring the crucial role of the EM iteration in the training process.




\section{Related Work}

The almost KGA methods use translation-based or GNN-based models and reflect relation, head entity, and tail entity into a uni-latent space. 
TransE \cite{bordes2013translating} is first proposed and performs well in one-to-one link prediction tasks.  
MTransE \cite{chen2017multilingual}  utilizes the TransE model to learn entity embeddings for two Knowledge Graphs separately and designed a space transformation mechanism for the alignment. 
BootEA \cite{sun2018bootstrapping} proposes an embedding-based bootstrapping method for entity alignment, which iteratively labels potential entity alignments as training data to learn alignment-guided KG embeddings. However, these methods  may fail to adequately
capture these complex structural characteristics.
The additional related work is listed in Appendix \ref{sec:appendix-rw}.

\section{Conclusion}
 EREM mainly consists of Entity Matching module and Relation Matching module and uses an EM optimization framework for multiple enhancement of the EA and RA. 
EREM alternatively updates the EA and RA via an E-step and an M-step.
In each step, both EA and RA are mutually enhanced by learning from the anchors predicted by the other module. 
Moreover, our work applies LLMs to EREM and designs a CoT strategy.
Extensive experiments conducted on EA and RA validation datasets demonstrate the effectiveness and efficiency of EREM.

\bibliography{ref}
\bibliographystyle{acl_natbib}

\appendix

\section{Limitations}
EREM is an Expectation-Maximization-based framework with two tasks: EA and RA. EREM can be adapted to most EA models. We have also applied it to LLMs and discussed the results. However, EREM heavily relies on the models it adapts to and is highly sensitive to them. 
EREM has relatively low time consumption but requires a significant amount of GPU resources. This could make it unsuitable for particularly large-scale datasets, such as those in the million-level range.
EREM don't consider the dangling entities and relations when do EA and RAa.
Finally,  EREM is unable to identify ambiguous relations for RA, which makes it hard to optimize RA.

\section{Integrating Large Language Mode Performance into EREM}
\label{sec:appendix-llm}

\begin{table*}[h]
\begin{tabularx}{\textwidth}{p{20mm} X p{25mm}}
\hline
Steps                              & Prompt template & LLM \\ \hline
Task Description                          &     You are a good assistant to perform entity alignment and relation alignment. I will give a question and a list of candidate answers to this question. You need to choose the best answer from the candidate list based on its given description information and your own knowledge. If no answer from the candidate list, please answer None.            &  Yes   \\ \hline
Initial Relation Align                    &     Given relation <RELATION\_0>, please choose same relation from the candidate list [“< RELATION\_1 >”, “< RELATION\_1 >”, …]. You must respond with one corresponding choice at most. If no answer from the candidate list, please answer None.            &  Output answer or None answer   \\ \hline
Initial Entity Align                      &   Given entity <ENTITY\_0>, please choose same entity from the candidate list [“<ENTITY\_1 >”, “<ENTITY\_2>”, …]. You must respond with one corresponding choice at most. If no answer from the candidate list, please answer None.              &  Output answer or None answer   \\ \hline
Description Entity based Aligned Relation &    
For <ENTITY\_0> contains triples: (ENTITY\_0, RELATION\_0, ENTITY\_1), …; For <ENTITY\_2> contains triples: (ENTITY\_2, RELATION\_1, ENTITY\_3), ...; <RELATION\_0> and< RELATION\_1> are the same relation.
             &   Okay, I understand. I will wait for your new query.  \\ \hline
Rethinking Entity Align                   &    Is the entity alignment pair (ENTITY\_0, ENTITY\_2) satisfactory enough? (YES or NO ). If response No, reselect  entity from entity candid list [“<ENTITY\_1 >”, “<ENTITY\_3>”, …].             &   YES/NO the true entity is < ENTITY >  \\ \hline
Description Relation based Aligned Entity &    
For < ENTITY\_0> contains triples: (ENTITY\_0, RELATION\_0, ENTITY\_1), …; For < ENTITY\_2> contains triples: (<ENTITY\_3>, <RELATION\_1>, <ENTITY\_2>), …;   < RELATION\_0> and< RELATION\_1> are the same relation.
           &  Okay, I understand. I will wait for your new query.   \\ \hline
Rethinking Relation Align       &    Is the relation alignment pair (RELATION\_0, RELATION\_1) satisfactory enough? (YES or NO ). If response No, reselect relation from relation candid list [“< RELATION\_2>”, “< RELATION\_3>”, …].             &  YES/NO the true relation is <RELATION >   \\ \hline
\end{tabularx}
\caption{ Templates of EREM for each step. <ENTITY> would be entity name and <RELAION> would be relation name.}
\label{tab:cot}
\end{table*}

In recent years, Large Language Models (LLMs) \cite{touvron2023llama, chang2024survey, kasneci2023chatgpt} have exhibited exceptional performance across a wide range of natural language processing applications.
Concurrently, several approaches \cite{li2024contextualization, bi2024codekgc} have investigated the capacity of LLMs to enhance KG. 
LLMEA \cite{yang2024two} innovatively applies LLMs to the task of selecting equivalent entities from candidate entities. This is achieved by utilizing the internal contextual knowledge of the LLMs through the implementation of a meticulously crafted knowledgeable prompt.
To harness the capabilities of KGE methods, candidates are generated based on the embedding similarity between the two knowledge graphs. Specifically, the top $k$ candidate entities with the highest similarity scores are selected.
Employing 1-shot In-Context Learning (ICL), ChatEA \cite{jiang2024unlocking} designs a KG-Code Translation framework to incorporate contextual information of entities, including neighboring entities, related relations, and temporal data, into entity description sentences. This approach facilitates the comprehension of these descriptions by LLMs.
However, these methods do not consider relation alignment as an independent procedure and consistently incorporate all contextual information of entities. 
This approach may introduce excessive additional context, making it challenging for LLMs to effectively utilize this information for accurate inference.

The Chain-of-Thought (CoT) \cite{wei2022chain} prompting strategy significantly enhanced the reasoning performance of LLMs without necessitating additional fine-tuning.
Recently, certain knowledge enhancement methods \cite{xu2024multi,  wei2023kicgpt} have leveraged CoT prompting techniques to unlock the knowledge capabilities inherent in LLMs. 
Building upon these methodologies, we propose a CoT method to integrate semantic knowledge from both our framework and LLMs, thereby unveiling a vast yet under-explored potential in KGA.
The proposed CoT prompting strategy enables LLMs to systematically deconstruct complex queries, guiding them step by step to provide accurate answers.
Before applying the CoT strategy, it's necessary to generate candidate entities and relations. We utilize LaBSE to create embeddings for entities and relations based on their names.
By computing the cosine similarity between the source embedding and target embedding, we can sort the computed similarities and select the top 10 entities and relations. These selections form the candidate list for EA and RA in the iteration.
Our CoT strategy to enhance KGA joint of EA and RA contains six main steps:

\begin{itemize}
    \item \textbf{Step 1. Initial Relation Align:} Similar to entity alignment, we treat relation alignment as a selection problem. This approach guides LLMs to choose the correct answer from the candidate relations. This process can be viewed as a cold start approach for RA, relying solely on the relation names. The step will output the relation alignment pairs.
    \item \textbf{Step 2. Initial Entity Align:} This step aims to initialize the entity alignment pairs, which are also framed as a selection problem. The output of this step will be the relation alignment pairs. Both step 1 and step 2 are executed only once.
    \item \textbf{Step 3. Description Entity based Aligned Relation:} Contrary to the exited entities description generation methods, given aligned entity pair ($e_1$, $e_2$) and aligned relation pair ($r_1$, $r_2$), if $e_1$ with context relation $r_1$ and $e_2$ with context relation $r_2$, the description information will be created by this triple contained these aligned relation and entity. The rethinking processing in step 4 will see this description information as a few-shot case.
    
    \item \textbf{Step 4. Rethinking Entity Align:}  For reasoning, through the in-context learning along with few-shot cases, the model will consider the correction of pre-aligned entity pairs in Step 2. If the rethink processing answer is "Yes", the alignment is considered satisfactory. Otherwise, the model will re-choose entities from the candidate list, which will  remove the unaligned entity after rethinking. The step 3 and 4 use RA to enhance EA.
    
    \item \textbf{Step 5. Description Relation based Aligned Entity:}  Same as Step 4, if $r_1$ with context entity $e_1$ and $r_2$ with context entity $e_2$, the description information will also be created by this triple containing these aligned relation and entity. 
    
    \item \textbf{Step 6. Rethinking Relation Align: } This step is designed as the same as step 4.  The step 5 and 6 use RA to enhance EA.
\end{itemize}

The step 3, 4, 5, and 6 will be iteratively executed in model training processing step by step.
The detailed prompt template for all steps can be seen in Table \ref{tab:cot}. To help LLM better understand our goal, we add a task description step before our CoT strategy.
Our exploration commences with a scrupulous evaluation of LLMs, such as ChatGLM-6B and online ChatGLM-3, in KGA.
We follow the previous LLM-based methods to assess the performance of directly instructing LLMs to perform EA and RA.
In more detail, we provide detailed examples to illustrate the input and output of each CoT step. We present examples in Table \ref{tab:cot-case-1}, \ref{tab:cot-case-2}, \ref{tab:cot-case-3}, \ref{tab:cot-case-4}, \ref{tab:cot-case-5}, and \ref{tab:cot-case-6}. In our experiments, ChatGLM6b with step 1 and step 2 can be seen as a foundation model named ChatGLM6b.

\begin{table}[htbp]
\begin{tabularx}{\linewidth}{p{15mm} X }
\hline
Steps                                    & Initial Relation Align  \\ \hline
Prompt                                    & \begin{CJK*}{UTF8}{gbsn} Given relation “国家”, please choose same relation from the candidate list  ['country', 'birthPlace', 'deathPlace', 'subdivisionName', 'headquarters', 'origin', 'leaderName', 'restingplace', 'house', 'burialPlace']. You must respond with one corresponding choice at most. If no answer from the candidate list, please answer None. \end{CJK*}  \\ \hline
LLM Output                                    & country \\ \hline
\end{tabularx}
\caption{ The case of Initial Relation Align step. }
\label{tab:cot-case-1}
\end{table}

\begin{table}[htbp]
\begin{tabularx}{\linewidth}{p{15mm} X }
\hline
Steps                                    & Initial Entity Align \\ \hline
Prompt                                    & \begin{CJK*}{UTF8}{gbsn} Given entity “杜兰大学”, please choose a same entity from the candidate list ["Durham University", "Tulane University", "University of Dundee", "Duke University", "Lund University", "DePaul University", "Brown University", "University of Delhi", "Auburn University", "Leiden University"]. You must respond with one corresponding choice at most. If no answer from the candidate list, please answer None.\end{CJK*}  \\ \hline
LLM Output                                    & Durham University \\ \hline
\end{tabularx}
\caption{ The case of Initial Entity Align step. }
\label{tab:cot-case-2}
\end{table}

\begin{table}[htbp]
\begin{tabularx}{\linewidth}{p{15mm} X }
\hline
Steps                                    & Description Entity based Aligned Relation \\ \hline
Prompt                                    & \begin{CJK*}{UTF8}{gbsn} For “杜兰大学”, contains triples: (“杜兰大学”, “国家”, “美国”)
For “Durham University”, contains triples: (“Durham University”, “country”, “United Kingdom”).
“国家” and “country” are the same relation.
\end{CJK*}  \\ \hline
LLM Output                                    & Okay, I understand. I will wait for your new query. \\ \hline
\end{tabularx}
\caption{ The case of Description Entity based Aligned Relation step. }
\label{tab:cot-case-3}
\end{table}

\begin{table}[htbp]
\begin{tabularx}{\linewidth}{p{15mm} X }
\hline
Steps                                    & Rethinking Entity Align \\ \hline
Prompt                                    & \begin{CJK*}{UTF8}{gbsn} Is the entity alignment pair (“杜兰大学”, “Durham University”) satisfactory enough? (YES or NO ). If response No, reselect entity from entity candid list ["Tulane University",  "University of Dundee", "Duke University", "Lund University", "DePaul University", "Brown University", "University of Delhi", "Auburn University", "Leiden University"].
\end{CJK*}  \\ \hline
LLM Output                                    & No, the true align entity is Tulane University. \\ \hline
\end{tabularx}
\caption{ The case of Rethinking Entity Align. }
\label{tab:cot-case-4}
\end{table}

\begin{table}[htbp]
\begin{tabularx}{\linewidth}{p{15mm} X }
\hline
Steps                                    & Description Relation based Aligned Entity \\ \hline
Prompt                                    & \begin{CJK*}{UTF8}{gbsn} For “国家”, contains triples: (“杜兰大学”, “国家”, “美国”)、(“耶鲁法学院”, “国家”, “美国”);
For “country”, contains triples: (“Tulane University”, “country”, “United States”)、(“Yale Law School”, “country”, “United States”);
“杜兰大学” and “Tulane University” are the same entity.
“耶鲁法学院” and “Yale Law School” are the same entity.
“美国” and “United States” are the same entity.
\end{CJK*}  \\ \hline
LLM Output                                    & Okay, I understand. I will wait for your new query. \\ \hline
\end{tabularx}
\caption{ The case of Description Relation based Aligned Entity. }
\label{tab:cot-case-5}
\end{table}

\begin{table}[ht]
\begin{tabularx}{\linewidth}{X X }
\hline

\multicolumn{2}{c}{EA case study} \\ \hline
\begin{CJK*}{UTF8}{gbsn}明成祖 \end{CJK*}       & Yongle Emperor        \\ \hline
\begin{CJK*}{UTF8}{gbsn}朱元璋 \end{CJK*}        & Hongwu Emperor           \\ \hline
\begin{CJK*}{UTF8}{gbsn}官话 \end{CJK*}        & Mandarin Chinese      \\ \hline
\multicolumn{2}{c}{RA case study} \\ \hline
\begin{CJK*}{UTF8}{gbsn}国际组织\end{CJK*}      & international         \\ \hline
\begin{CJK*}{UTF8}{gbsn}拥有者  \end{CJK*}     & owner                 \\ \hline
\begin{CJK*}{UTF8}{gbsn}货币单位\end{CJK*}      & currency              \\ \hline
\end{tabularx}
\caption{The Case Study Table for EREM.}
\label{case-study}
\end{table}

\begin{table}[htbp]
\begin{tabularx}{\linewidth}{p{15mm} X }
\hline
Steps                                    & Rethinking Relation Align \\ \hline
Prompt                                    & \begin{CJK*}{UTF8}{gbsn} Is the relation alignment pair (“国家”, “country”) satisfactory enough? (YES or NO ). If response No, reselect relation from relation candid list [ 'birthPlace', 'deathPlace', 'subdivisionName', 'headquarters', 'origin', 'leaderName', 'restingplace', 'house', 'burialPlace'].
\end{CJK*}  \\ \hline
LLM Output                                    & Yes. \\ \hline
\end{tabularx}
\caption{ The case of Rethinking Relation Align. }
\label{tab:cot-case-6}
\end{table}

\section{Case Study}
\label{sec:casestudy}

Figure \ref{case-study} shows an example in  DBP15K-ZH\_EN and the entity linked pairs and relation linked pairs are not aligned by FGWEA. After adopted our EREM, this incorrectly aligned entities and relations will be correctly aligned. Considering the mutual enhancement among EA and RA, EREM will achieve a good performance.

\section{Notation and Explanation}
\label{sec:notaion}

In this section, we summarize the notations in Table \ref{tab:notation}.

\begin{table}[htbp]
\begin{tabularx}{\linewidth}{p{15mm} X }
\hline
Notation & Explanation \\ \hline
         G&             Knowledge graph\\
         $\mathcal{E}$&             Entity set in KG\\
 $\mathcal{R}$&Relation set in KG\\
         $e_i$&             Entity $i$ from source $G$\\
         $e_i^{\prime}$&             Entity $i$ from target $G'$\\ 
 $r_j$&Relation $j$ from source $G$\\
 $r_j^{\prime}$&Relation $j$ from target $G'$\\
 $h$&Entity or relation embedding\\
 $g$&Embedding module\\
 $C^{e}$&Entity cost matrix\\
 $C^{ent}$&Entity embedding matrix \\
 $ C^{stru}$&Entity structure-aware matrix \\
 $C^{rel}$& Relation-aware matrix\\
 $C^{r}$&Relation cost matrix\\
 $C^{rel}$&Relation embedding matrix\\
 $C^{stru\_rel}$&Relation structure similarity matrix\\
 $\Psi^e$&Entity transport matrix\\
 $\Psi^r$&Relation transport matrix\\
 $ \overline{y}_e $& Entity anchor\\
 $ \widehat{y_e} $&Hard entity anchor\\
 $ \overline{y}_r $&Relation anchor\\
 $ \widehat{y_r} $&Hard relation anchor\\ \hline
\end{tabularx}
\caption{Notation summary}
\label{tab:notation}
\end{table}

\section{Supplementary Experiment}
\label{sec:appendix}

\begin{table}[htbp]
\begin{tabularx}{\linewidth}{p{29mm}XXX}
\hline
Model         & ZH\_EN & JA\_EN & FR\_EN \\ \hline
GCNAlign      & 23min            & 24min            & 25min            \\
GCNAlign-EREM & 47min            & 46min            & 47min            \\ \hline
BootEA        & 492min           & 591min           & 783min           \\
BootEA-EREM   & 513min           & 609min           & 801min           \\ \hline
\end{tabularx}
\caption{Time cost comparison.}
\label{tab:time-cost}
\end{table}

\textbf{Time Cost Comparison}. We specifically evaluate training convergence time costs of some methods and EREM in Table \ref{tab:time-cost}. 
The efficiency of EREM is high, as it can achieve significant improvements over the original model with just a relatively short training time.
Typically, 
BootEA-EREM only needs an additional 21 minutes of training to improve the model’s performance by 28\% in DBP15K-ZH\_EN, which uses a bit of time and achieves a great promotion compared with BootEA long training time.

\begin{figure*}[h]
    \begin{minipage}[t]{0.33\linewidth}
        \centering
        \includegraphics[scale=0.17]{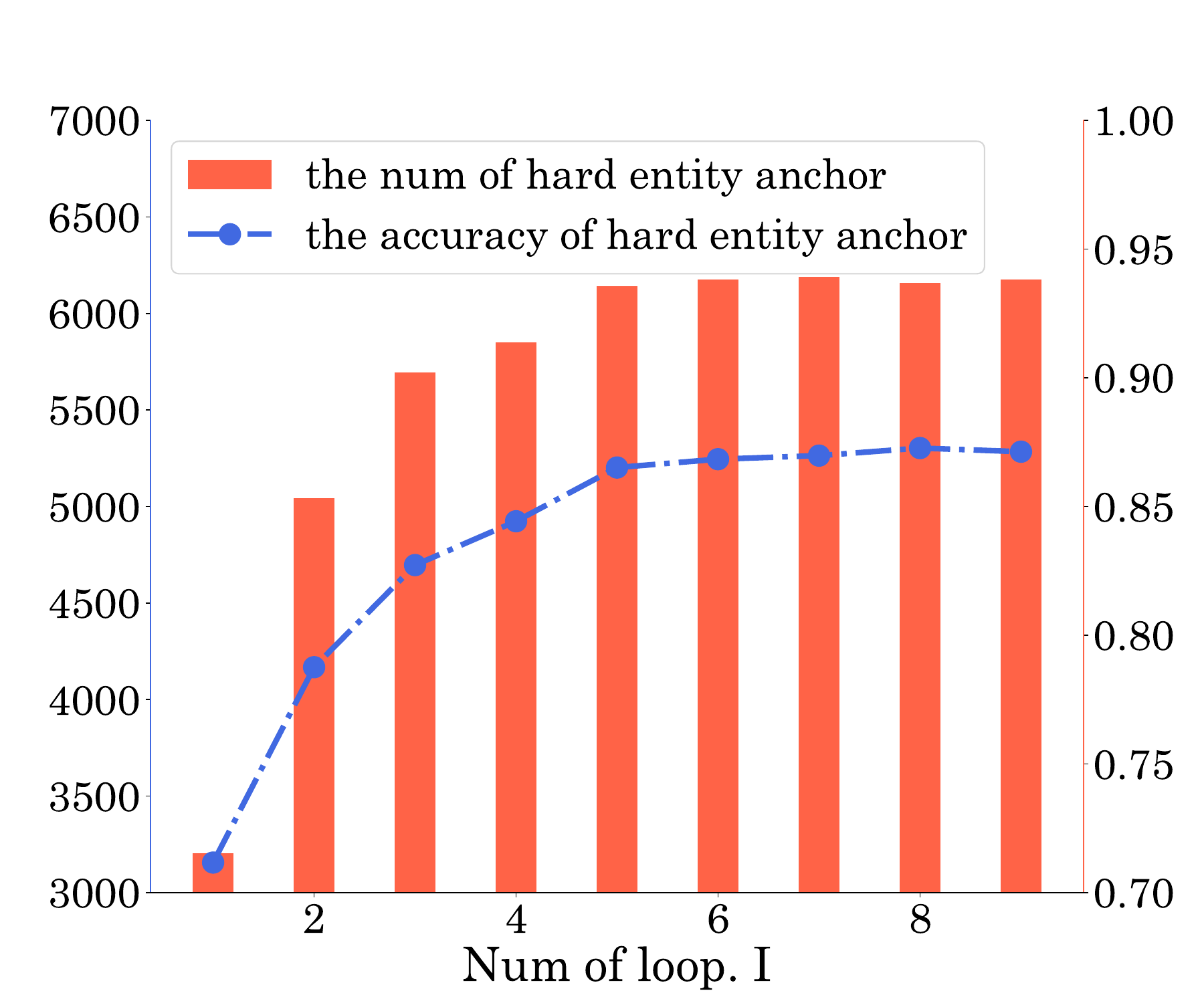}
        \label{fig:1}
    \end{minipage}%
    \begin{minipage}[t]{0.33\linewidth}
        \centering
        \includegraphics[scale=0.17]{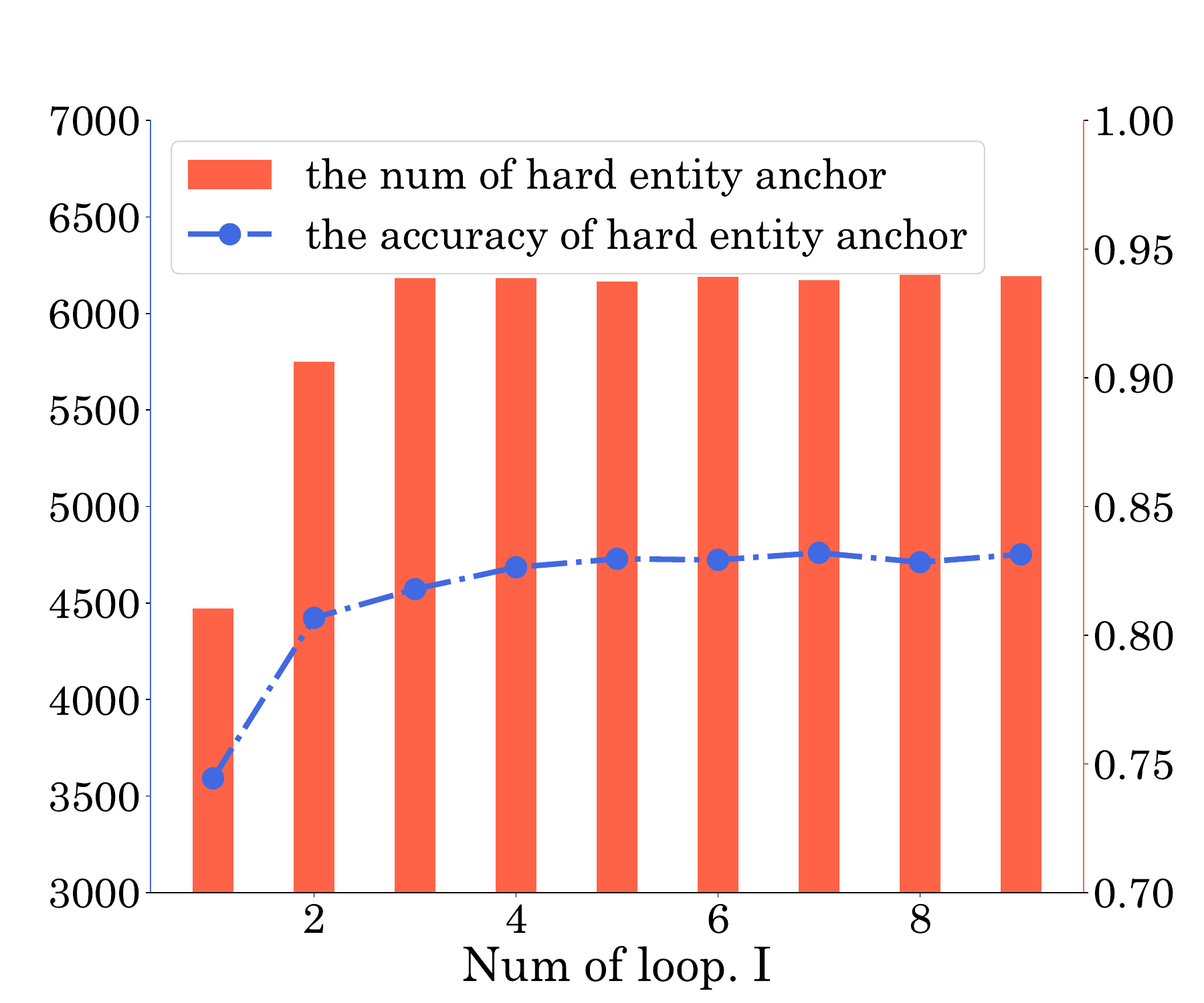}
        \label{fig:2}
    \end{minipage}
    \begin{minipage}[t]{0.33\linewidth}
        \centering
        \includegraphics[scale=0.17]{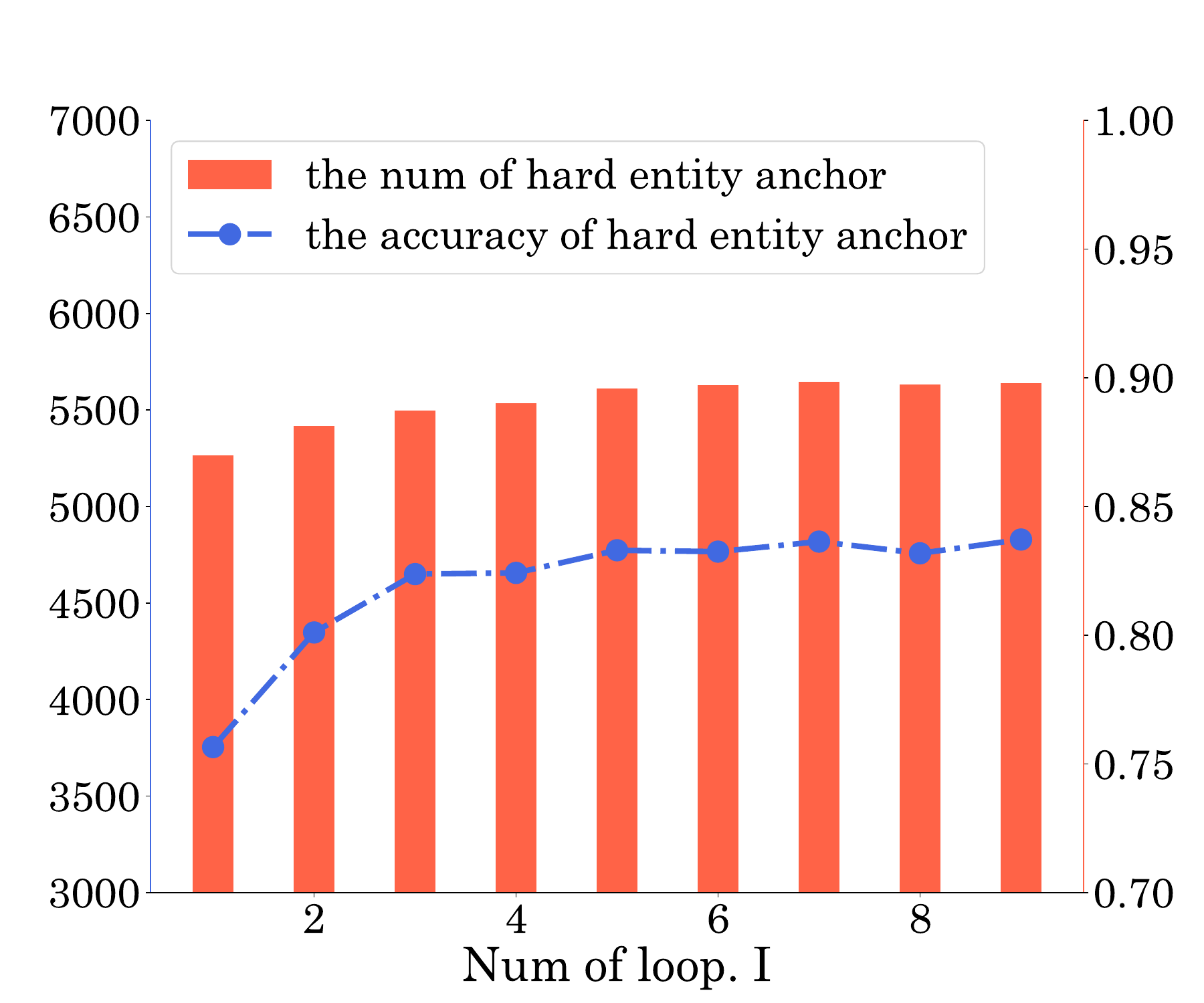}
        \label{fig:3}
    \end{minipage}
    \caption{ 
The three sets of figures illustrate the rise in accuracy and quantity of hard entity anchors during the iterative process (from iter=1 to 9), demonstrating the repair effect of relation alignment on entity alignment. (The data is extracted when the embedding module is GCN-Align.) 
}
     \label{fig:hard_anchor}
\end{figure*}

\begin{figure}[htbp]
        \centering
        \includegraphics[scale=0.34]{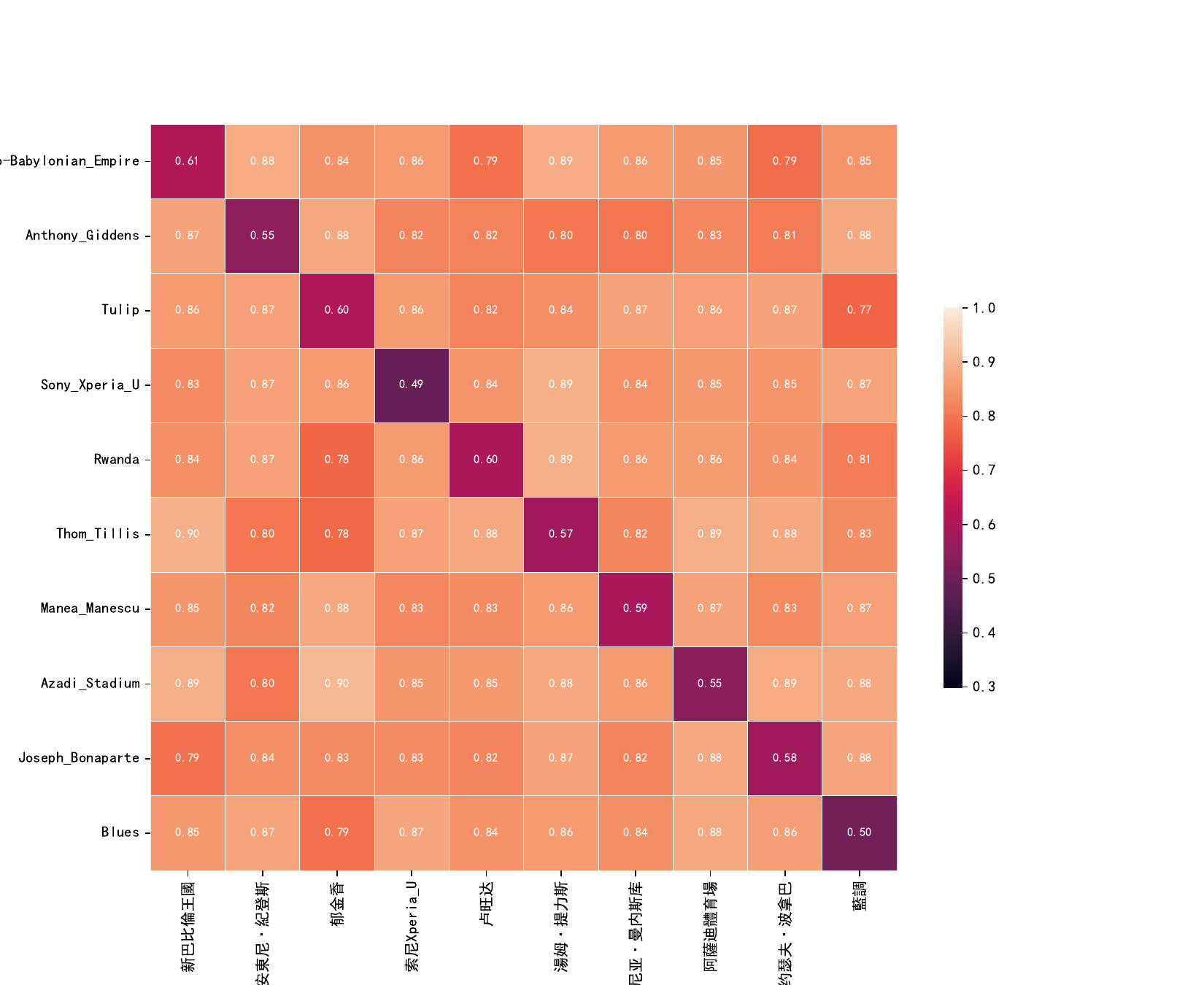}
        \caption{The entity alignment cost matrix before award.}
        \label{fig:hot_1}

\end{figure}

\begin{figure}[htbp]

        \centering
        \includegraphics[scale=0.34]{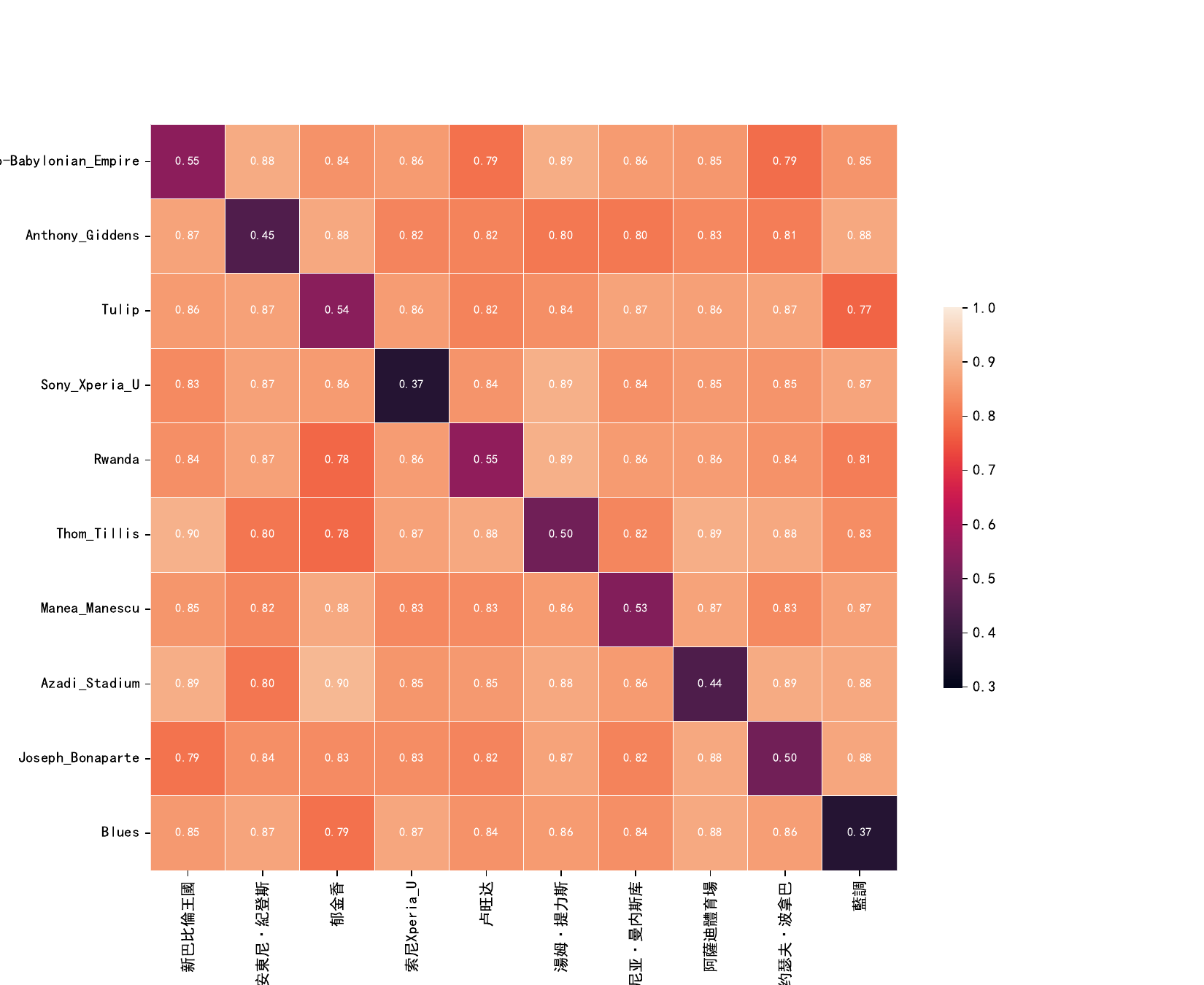}
        \caption{ The entity alignment cost matrix after award.}
        \label{fig:hot_2}

\end{figure}

\section{Additional Related Work}
\label{sec:appendix-rw}

In recent years, to address how to use cross-KG relation information, recent research  has been dedicated to relation-aware mechanisms to explore complex and heterogeneous relation information.  
The ICMEA \cite{zeng2022interactive} solves shortcomings in SelfKG and designs a Relation-aware neighborhood aggregator, intending to update entity embeddings by executing message passing using the relation information present in the KG. 
The RNM \cite{zhu2021relation} tries to explore useful information from the connected relations, when do entity neighborhood matching.
In the RAEA \cite{zhu2023cross}, the representation of relations is integrated into entity representation using the Dual-Primal Graph Convolutional Neural Network (DPGCNN), and structure representation and attribute representation are learned by GCNs. 
The RAGA \cite{zhu2021raga} framework, based on Relation-aware Graph Attention Networks, captures interactions between entities and relations by employing a self-attention mechanism. This mechanism spreads entity information to relations and subsequently aggregates relation information back to entities.
FGWEA \cite{tang2023fused} utilizes the Fused Gromov-Wasserstein  distance to devise a three-stage progressive optimization algorithm. This approach considers cross-KG structural and relational consistencies within the optimization objectives.

\end{document}